\documentclass{article} 
\usepackage{iclr2024_conference,times}

\usepackage{amsmath,amsfonts,bm}

\def\eqref#1{equation~\ref{#1}}

\def\1{\bm{1}}

\DeclareMathAlphabet{\mathsfit}{\encodingdefault}{\sfdefault}{m}{sl}
\SetMathAlphabet{\mathsfit}{bold}{\encodingdefault}{\sfdefault}{bx}{n}

\usepackage{hyperref}
\usepackage{url}

\usepackage{graphicx}
\usepackage{multirow}
\usepackage{mathabx,graphicx}
\usepackage{makecell}
\usepackage{booktabs}
\usepackage{color}
\usepackage{colortbl}
\usepackage{multirow}
\usepackage{wrapfig}

\newcommand\Tstrut{\rule{0pt}{2.3ex}}         

\title{LRM: Large Reconstruction Model for \\Single Image to 3D}

\author{Yicong Hong$^{1,2}$\thanks{Intern at Adobe Research.} \quad Kai Zhang$^{1}$ \quad Jiuxiang Gu$^{1}$ \quad Sai Bi$^{1}$ \quad \textbf{Yang Zhou}$^{1}$ \\ 
\vspace{2pt}
\textbf{Difan Liu}$^{1}$ \quad \textbf{Feng Liu}$^{1}$ \quad \textbf{Kalyan Sunkavalli}$^{1}$ \quad \textbf{Trung Bui}$^{1}$ \quad \textbf{Hao Tan}$^{1}$ \\
\vspace{2pt}
$^{1}$Adobe Research \quad $^{2}$Australian National Univeristy \\
\texttt{mr.yiconghong@gmail.com} \\
\texttt{\{kaiz,jigu,sbi,yazhou,diliu,fengl,sunkaval,bui,hatan\}@adobe.com} \\
}

\newcommand{\ours}{LRM}

\iclrfinalcopy 
\begin{document}

\maketitle

\begin{abstract}

We propose the first Large Reconstruction Model (\ours{}) that predicts the 3D model of an object from a single input image within just 5 seconds.
In contrast to many previous methods that are trained on small-scale datasets such as ShapeNet in a category-specific fashion, \ours{} adopts a highly scalable transformer-based architecture with 500 million learnable parameters to directly predict a neural radiance field (NeRF) from the input image. We train our model in an end-to-end manner on massive multi-view data containing around 1 million objects, including both synthetic renderings from Objaverse and real captures from MVImgNet. This combination of a high-capacity model and large-scale training data empowers our model to be highly generalizable and produce high-quality 3D reconstructions from various testing inputs, including real-world in-the-wild captures and images created by generative models. Video demos and interactable 3D meshes can be found on our LRM project webpage: \textcolor{red}{\tt\small\url{https://yiconghong.me/LRM}}.
 
\end{abstract}

\section{Introduction}

Imagine if we could instantly create a 3D shape from a single image of an arbitrary object. Broad applications in industrial design, animation, gaming, and AR/VR have strongly motivated relevant research in seeking a generic and efficient approach towards this long-standing goal. 
Due to the underlying ambiguity of 3D geometry in a single view, early learning-based methods usually perform well on specific categories, utilizing the category data prior to infer the overall shape~\citep{yu2021pixelnerf}.
Recently, advances in image generation, such as DALL-E~\citep{ramesh2021dalle} and Stable Diffusion~\citep{rombach2022stablediffuse}, have inspired research that leverages the remarkable generalization capability of 2D diffusion models to enable multi-view supervision~\citep{liu2023zero123,tang2023makeit3d}. However, many of these methods require delicate parameter tuning and regularization, and their results are limited by the pre-trained 2D generative models. Meanwhile, there are many approaches that rely on per-shape optimization (\textit{e.g.} optimize a NeRF~\citep{mildenhall2021nerf,chan2022eg3d,Chen2022ECCV,mueller2022instant,SunSC22}) to construct a consistent geometry; this process is often slow and impractical.

On the other hand, the great success in natural language processing~\citep{devlin2018bert,brown2020gpt3,chowdhery2022palm} and image processing~\citep{caron2021dino,radford2021clip,alayrac2022flamingo,ramesh2022dalle2} can be largely credited to three critical factors: (1) using highly scalable and effective neural networks, such as the Transformers~\citep{vaswani2017attention}, for modeling the data distribution, (2) enormous datasets for learning generic priors, as well as (3) self-supervised-like training objectives that encourage the model to discover the underlying data structure while maintaining high scalability. 
For instance, the GPT (generative pre-trained transformer) series~\citep{radford2019language, brown2020gpt3, openai2023gpt4} build large language models with huge transformer networks, large-scale data, and the simple next-word prediction task.
In light of this, we pose the same question for 3D: given sufficient 3D data and a large-scale training framework, \textbf{\textit{is it possible to learn a generic 3D prior for reconstructing an object from a single image?}}

In this paper, we propose a \textbf{L}arge \textbf{R}econstruction \textbf{M}odel (\ours{}) for single-image to 3D. Our method adopts a large transformer-based encoder-decoder architecture for learning 3D representations of objects from a single image in a data-driven manner. Our method takes an image as input and regresses a NeRF in the form of a triplane representation~\citep{chan2022eg3d}. Specifically, \ours{} utilizes the pre-trained visual transformer DINO~\citep{caron2021dino} as the image encoder to generate the image features, and learns an image-to-triplane transformer decoder to project the 2D image features onto the 3D triplane via cross-attention and model the relations among the spatially-structured triplane tokens via self-attention.
The output tokens from the decoder are reshaped and upsampled to the final triplane feature maps. 
Afterwards, we can render the images at an arbitrary view by decoding the triplane feature of each point with an additional shared multi-layer perception (MLP) to get its color and density and performing volume rendering.

The overall design of \ours{} maintains high scalability and efficiency. In addition to the use of a fully transformer-based pipeline, a triplane NeRF is a concise and scalable 3D representation since it is computationally friendly compared to other representations such as volumes and point clouds. It also has a better locality with respect to the image input compared to tokenizing the NeRF’s model weights as in Shap-E~\citep{jun2023shape}. Moreover, our \ours{} is trained by simply minimizing the difference between the rendered images and ground truth images at novel views, without excessive 3D-aware regularization or delicate hyper-parameter tuning, allowing the model to be very efficient in training and adaptable to a wide range of multi-view image datasets.

To the best of our knowledge, \ours{} is the first \emph{large-scale 3D reconstruction model}; it contains more than 500 million learnable parameters, and it is trained on approximately one million 3D shapes and video data across diverse categories~\citep{deitke2023objaverse,yu2023mvimgnet}; this is substantially larger than recent methods that apply relatively shallower networks and smaller datasets~\citep{chang2015shapenet,reizenstein2021co3d,downs2022googlescanned}.
Through experiments, we show that \ours{} can reconstruct high-fidelity 3D shapes from a wide range of images captured in the real world, as well as images created by generative models. \ours{} is also a highly practical solution for downstream applications since it can produce a 3D shape in just five seconds\footnote{Five seconds per shape on a single NVIDIA A100 GPU, including around 1.14 seconds image-to-triplane feed-forward time, 1.14 seconds to query resolution of $384{\times}384{\times}384$ points from the triplane-NeRF, and 1.91 seconds mesh extraction time using Marching Cubes~\citep{lorensen1998marchingcubes}.} without post-optimization.

\section{Related work}
\label{sec:relatedwork}

\paragraph{Single Image to 3D Reconstruction}
Extensive efforts have been devoted to address this problem, including early learning-based methods that explore point clouds~\citep{fan2017pointset,wu2020pq}, voxels~\citep{choy20163dr2d2,tulsiani2017drc,chen2019imnet}, and meshes~\citep{wang2018pixel2mesh,gkioxari2019meshrcnn}, as well as various approaches that learn implicit representations such as SDFs~\citep{park2019deepsdf,mittal2022autosdf}, occupancy networks~\citep{mescheder2019occupancy}, and NeRF~\citep{jang2021codenerf,muller2022autorf}.
Leveraging 3D templates~\citep{roth2016adaptive,goel2020without,kanazawa2018learning,kulkarni2020articulation}, semantics~\citep{li2020self}, and poses~\citep{bogo2016keepsmpl,novotny2019c3dpo} as shape priors have also been widely studied in category-specific reconstruction. Category-agnostic methods show great generalization potential~\citep{yan2016perspective,niemeyer2020differentiable}, but they often unable to produce fine-grained details even when exploiting spatially-aligned local image features~\citep{xu2019disn,yu2021pixelnerf}.

Very recently, there is an emerging trend of using pre-trained image/language models~\citep{radford2021clip,li2022blip,li2023blip2,saharia2022imagen,rombach2022stablediffuse}, to introduce semantics and multi-view guidance for image-to-3D reconstruction~\citep{liu2023zero123,tang2023makeit3d,deng2023nerdi,shen2023anything3d,anciukevivcius2023renderdiffusion,melas2023realfusion,metzer2023latentnerf,xu2023neurallift360,qian2023magic123,li2023instant3d}. For instance, Zero-1-to-3 fine-tunes the Stable Diffusion model to generate novel views by conditioning on the input image and camera poses~\citep{liu2023zero123}; its view consistency and reconstruction efficiency have been further improved by~\cite{liu2023one2345}. Make-It-3D~\citep{tang2023makeit3d} uses BLIP to generate text descriptions for the input image (which is applied to guide the text-to-image diffusion) and trains the model with score distillation sampling loss~\citep{poole2022dreamfusion} and CLIP image loss to create geometrically and semantically plausible shapes.

In contrast to all these methods, our \ours{} is a purely data-driven approach that learns to reconstruct arbitrary objects in the wild. 
It is trained with minimal and extensible 3D supervision (\textit{i.e.}, rendered or captured 2D images of 3D objects) and does not rely on any guidance from pre-trained vision-language contrastive or generative models.

\paragraph{Learning 3D Representations from Images}
3D reconstruction from a single image is an ill-posed problem that has been frequently addressed by models with generative properties. Many previous works apply an encoder-decoder framework to model the image-to-3D data distribution~\citep{choy20163dr2d2,yan2016perspective,dai2017shape,xu2019disn,wu2020pq,muller2022autorf,sajjadi2022srt,goel2023humans4d}, where a compact latent code is trained to carry the texture, geometry, and pose details of the target. However, learning such an expressive representation usually requires a capable network and abundant 3D data which is very expensive to acquire. Hence most of these methods only focus on a few categories and produce very coarse results. 
GINA-3D~\citep{shen2023gina3d} implements a model that applies a visual transformer encoder and cross-attention (instead of a transformer decoder as in \ours{}) to translate images to triplane representations. However, the model and training are much smaller in scale, and their work has a different focus on category-specific generation.
Recent data-driven approach MCC~\citep{wu2023mcc} trains a generalizable transformer-based decoder with CO3D-v2 data~\citep{reizenstein2021co3d} to predict occupancy and color from the input image and its unprojected point cloud. Although MCC can handle real and generated images and scenes, the results are usually over-smooth and lose details.

\paragraph{Multimodal 3D} 

Motivated by the great advances in 2D multimodal learning~\citep{tan2019lxmert,chen2020uniter,chen2022visualgpt,yu2022coca,singh2022flava,wang2022internvideo,alayrac2022flamingo,girdhar2023imagebind}, \ours{} considers 3D as a new modality and directly grounds 2D feature maps onto 3D triplane via cross-attention. There are early attempts in this direction that minimize the difference between encoded image and 3D representations~\citep{girdhar2016learning,mandikal20183dlmnet}, as well as recent research, ULIP~\citep{xue2023ulip} and CLIP$^2$~\citep{zeng2023clip2}, which bridges 3D, language, and images via contrastive learning. LERF~\citep{kerr2023lerf} learns a language field inside NeRF by rendering CLIP embeddings along training rays.
In contrast, our method focuses on generic single image-to-3D reconstruction. 
We would like to mention the concurrent work Cap3D~\citep{luo2023cap3d} that produces descriptions for 3D shapes by applying BLIP~\citep{li2023blip2} to generate captions of different views, uses GPT-4~\citep{openai2023gpt4} to summarize them, and then employs these language-3D pairs for training text-to-3D generative models~\citep{nichol2022pointe,poole2022dreamfusion,jun2023shape}. 
There are also recent works in connecting 3D and large language models, such as 3D-LLM~\citep{hong20233dllm} and LLM-Grounder~\citep{yang2023llmgrounder}.

\begin{figure}[t]
  \centering
  \includegraphics[width=0.99\textwidth]{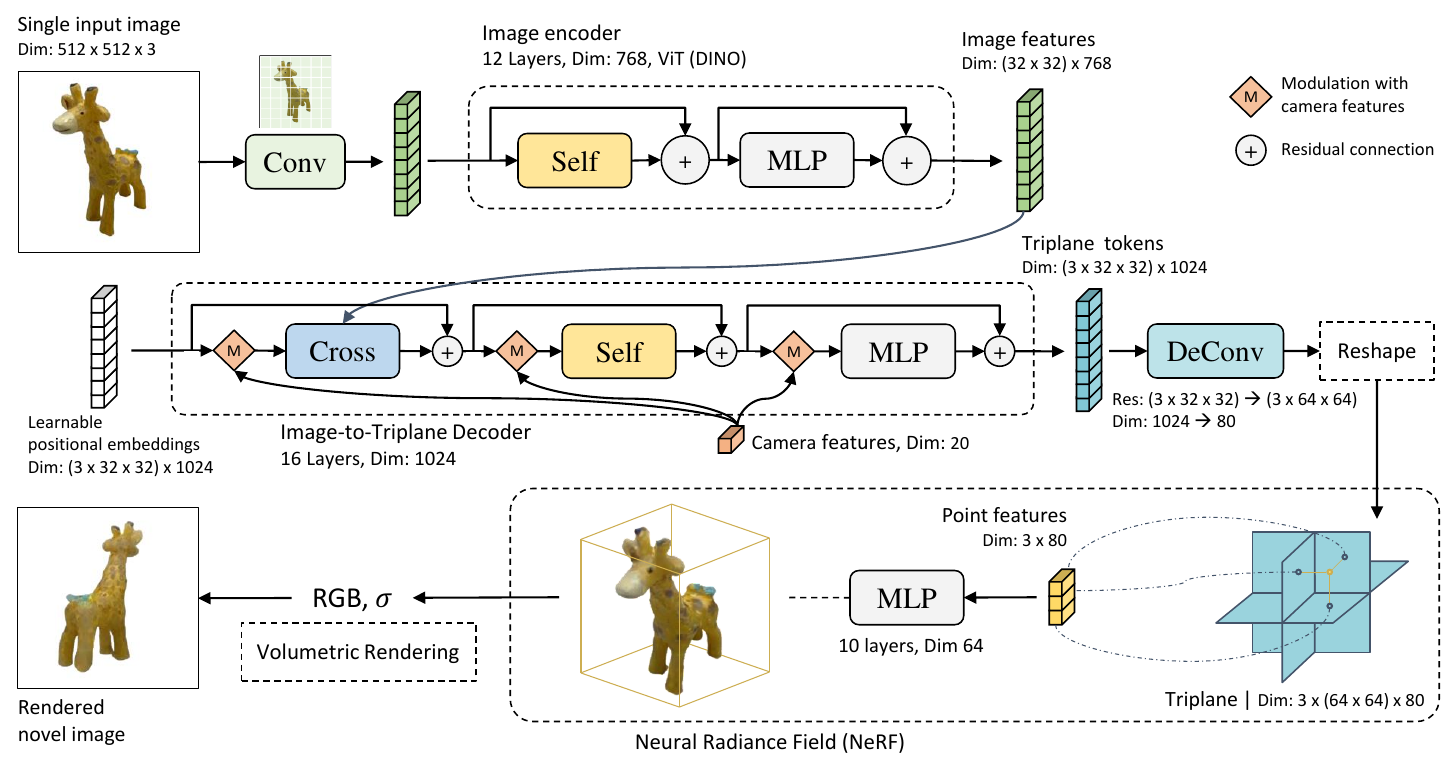}
  \vspace{-5pt}
  \caption{The overall architecture of \ours{}, a fully-differentiable transformer-based encoder-decoder framework for single-image to NeRF reconstruction. \ours{} applies a pre-trained vision model (DINO) to encode the input image (Sec.~\ref{subsec:encoder}), where the image features are projected to a 3D triplane representation by a large transformer decoder via cross-attention (Sec.~\ref{subsec:decoder}), followed by a multi-layer perceptron to predict the point color and density for volumetric rendering (Sec.~\ref{subsec:nerf}). The entire network is trained end-to-end on around a million of 3D data (Sec.~\ref{subsec:data}) with simple image reconstruction losses (Sec.~\ref{subsec:losses}). 
  }
  \label{fig:model}
\end{figure}

\section{Method}

In this section, we detail the proposed \ours{} architecture (Fig.~\ref{fig:model}). 
\ours{} contains an image encoder that encodes the input image to patch-wise feature tokens (Sec.~\ref{subsec:encoder}), followed by an image-to-triplane decoder that projects image features onto triplane tokens via cross-attention (Sec.~\ref{subsec:decoder}). 
The output triplane tokens are upsampled and reshaped into the final triplane representation, which is used to query 3D point features. 
Lastly, the 3D point features are passed to a multi-layer perception to predict RGB and density for volumetric rendering (Sec.~\ref{subsec:nerf}).
The training objectives and data are described in Sec.~\ref{subsec:losses} and Sec.~\ref{subsec:data}.

\subsection{Image Encoder} 
\label{subsec:encoder}

Given an RGB image as input, \ours{} first applies a pre-trained visual transformer (ViT)~\citep{dosovitskiy2020vit} to encode the image to patch-wise feature tokens $\{\boldsymbol{h}_{i}\}^{n}_{i=1}\in{\mathbb{R}^{d_E}}$, where $i$ denotes the $i$-th image patch, $n$ is the total number of patches, and $d_E$ is the latent dimension of the encoder. 
Specifically, we use DINO~\citep{caron2021dino}, a model trained with self-distillation that learns interpretable attention over the structure and texture of the salient content in images.  
Compared to other semantic-oriented representations such as the visual features from ImageNet-pretrained ResNet~\citep{he2016resnet} or CLIP~\citep{radford2021clip}, the detailed structural and texture information in DINO is more important in our case since \ours{} can use it to reconstruct the geometry and color in 3D space. 
As a result, instead of only using the ViT pre-defined class token \texttt{[CLS]} that aggregates patch-wise features, we also utilize the entire feature sequence $\{\boldsymbol{h}_{i}\}^{n}_{i=1}$ to better preserve this information\footnote{For simplicity, we use $\{\boldsymbol{h}_{i}\}^{n}_{i=1}$ in the following to denote the concatenated sequence of the encoded \texttt{[CLS]} token and patch-wise features.}.

\subsection{Image-to-Triplane Decoder}
\label{subsec:decoder}
We implement a transformer decoder to project image and camera features onto learnable spatial-positional embeddings and translate them to triplane representations. 
This decoder can be considered as a prior network that is trained with large-scale data to provide necessary geometric and appearance information to compensate for the ambiguities of single-image reconstruction.

\paragraph{Camera Features}
We construct the camera feature $\boldsymbol{c}\in\mathbb{R}^{20}$ of the input image by flattening out the 4-by-4 camera extrinsic matrix $\boldsymbol{E}$ (that represents the camera-to-world transformation) and concatenate it with the camera focal length $\mathit{foc}$ and principal point $\mathit{pp}$ as $\boldsymbol{c}=[\boldsymbol{E}_{1{\times}16}, \mathit{foc}_{x}, \mathit{foc}_{y}, \mathit{pp}_{x}, \mathit{pp}_{y} ]$.
Moreover, we normalize the camera extrinsic $\boldsymbol{E}$ by similarity transformations so that all the input cameras are aligned on the same axis (with the lookup direction aligned with the $z$-axis). 
Note that, \ours{} does not depend on a canonical pose of the object, and the ground truth $\boldsymbol{c}$ is only applied in training. 
Conditioning on normalized camera parameters greatly reduces the optimization space of triplane features and facilitates model convergence (see details in Sec.~\ref{subsec:implement}).
To embed the camera feature, we further implement a multi-layer perceptron (MLP) to map the camera feature to a high-dimensional camera embedding $\boldsymbol{\tilde{c}}$.
The intrinsics (focal and principal point) are normalized by the image's height and width before sending to the MLP layer.

\paragraph{Triplane Representation}
We follow previous works~\citep{chan2022eg3d,gao2022get3d} to apply triplane as a compact and expressive feature representation of the reconstruction subject. 
A triplane $\boldsymbol{T}$ contains three axis-aligned feature planes $\boldsymbol{T}_{X\!Y}$, $\boldsymbol{T}_{Y\!Z}$ and $\boldsymbol{T}_{X\!Z}$. 
In our implementation, each plane is of dimension $(64{\times}64){\times}{d_{T}}$ where $64 \times 64$ is the spatial resolution, and $d_T$ is the number of feature channels. 
For any 3D point in the NeRF object bounding box $[-1,1]^3$, we can project it onto each of the planes and query the corresponding point features $(\boldsymbol{T}_{{xy}},\boldsymbol{T}_{{yz}},\boldsymbol{T}_{{xz}})$
via bilinear interpolation, which is then decoded by an $\mathrm{MLP}^\mathit{nerf}$ into the NeRF color and density (Sec.~\ref{subsec:nerf}).

To obtain the triplane representation $\boldsymbol{T}$, 
we define learnable spatial-positional embeddings $\boldsymbol{f}^\mathit{init}$ of dimension $(3{\times}32{\times}32){\times}{d_D}$ which guide the image-to-3D projection and are used to query the image features via cross-attention, where $d_D$ is the hidden dimension of the transformer decoder.
The number of tokens in $\boldsymbol{f}^\mathit{init}$ is smaller than the number of final triplane tokens ($3{\times}64{\times}64$); we will upsample the output of the transformer $\boldsymbol{f}^\mathit{out}$ to the final $\boldsymbol{T}$.
In the forward pass, conditioning on the camera features $\boldsymbol{\tilde{c}}$ and image features $\{\boldsymbol{h}_{i}\}^{n}_{i=1}$, each layer of our image-to-triplane transformer decoder gradually updates the initial positional embedding $\boldsymbol{f}^\mathit{init}$ to the final triplane features via modulation and cross-attention, respectively. 
The reason for applying two different conditional operations is that the camera controls the orientation and distortion of the whole shape, whereas the image features carry the fine-grained geometric and color information that need to be embedded onto the triplane. 
Details of the two operations are explained below.

\paragraph{Modulation with Camera Features}
Our camera modulation is inspired by DiT~\citep{peebles2022dit} which implements an adaptive layer norm (adaLN) to modulate image latents with denoising timesteps and class labels.
Suppose $\{\boldsymbol{f}_j\}$ is a sequence of vectors in transformer, we define our modulation function $\mathrm{ModLN_{\boldsymbol{c}}}(\boldsymbol{f}_j)$ with camera feature $\boldsymbol{c}$ as
\begin{align}
\gamma, \beta &= \mathrm{MLP^{mod}}(\boldsymbol{\tilde{c}}) \\
\mathrm{ModLN_{c}}(\boldsymbol{f}_j) &= \mathrm{LN}(\boldsymbol{f}_j)\cdot(1+\gamma) + \beta
\end{align}
where $\gamma$ and $\beta$ are the scale and shift~\citep{huang2017arbitrary} output by $\mathrm{MLP^{mod}}$ and $\mathrm{LN}$ is the Layer Normalization~\citep{ba2016layernorm}. 
Such modulation is applied to each attention sub-layer which will be specified next.

\paragraph{Transformer Layers}
Each transformer layer contains a cross-attention sub-layer, a self-attention sub-layer, and a multi-layer perceptron sub-layer (MLP), where the input tokens to each sub-layer are modulated by the camera features.
Suppose feature sequence $\boldsymbol{f}^\mathit{in}$ is the input of an transformer layer, 
we can consider $\boldsymbol{f}^\mathit{in}$ as the triplane hidden features since they are corresponding to the final triplane features $\boldsymbol{T}$.
As shown in the decoder part of Fig.~\ref{fig:model}, the cross-attention module firstly attends from the triplane hidden features $\boldsymbol{f}^\mathit{in}$ to the image features $\{\boldsymbol{h}_{i}\}^{n}_{i=1}$, which can help linking image information to the triplane. 
Note that we here do not explicitly define any spatial alignment between the 2D images and 3D triplane hidden features, but consider 3D as an independent modality and ask the model to learn the 2D-to-3D correspondence by itself. 
The updated triplane hidden features will be passed to a self-attention sub-layer that further models the intra-modal relationships across the spatially-structured triplane entries. 
Then, a multi-layer perceptron sub-layer ($\mathrm{MLP}^\mathit{tfm}$) follows as in the original Transformer~\citep{vaswani2017attention} design.
Lastly, the output triplane features $\boldsymbol{f}^{out}$ will become the input to the next transformer layer. 

Such a design is similar to the Perceiver network~\citep{jaegle2021perceiver} while our model maintains a high-dimensional representation across the attention layers instead of projecting the input to a latent bottleneck. 
Overall, we can express this process for each $j$-th triplane entry in each layer as 
\begin{align}
    \boldsymbol{f}^\mathit{cross}_j &= \mathrm{CrossAttn}(\mathrm{ModLN}_{c}(\boldsymbol{f}^\mathit{in}_{j}); \{\boldsymbol{h}_{i}\}^{n}_{i=1}) + \boldsymbol{f}^\mathit{in}_j \\
    \boldsymbol{f}^\mathit{self}_j &= \mathrm{SelfAttn}(\mathrm{ModLN}_{c}(\boldsymbol{f}^\mathit{cross}_j); \{\mathrm{ModLN}_{c}(\boldsymbol{f}^\mathit{cross}_{j})\}_{j}) + \boldsymbol{f}^\mathit{cross}_j \\
    \boldsymbol{f}^\mathit{out}_j &= \mathrm{MLP^\mathit{tfm}}(\mathrm{ModLN}_{c}(\boldsymbol{f}^\mathit{self}_j)) + \boldsymbol{f}^\mathit{self}_j
\end{align} 
The $\mathrm{ModLN}$ operators in sub-layers (\textit{i.e.}, $\mathrm{CrossAttn}$, $\mathrm{SelfAttn}$, $\mathrm{MLP}^\mathit{tfm}$) use different set of learnable parameters in the layer normalization and the modulation $\mathrm{MLP}^\mathit{mod}$. 
We do not add additional superscript to differentiate them for clarity.

The transformer layers are processed sequentially.
After all the transformer layers, we obtain the output triplane features $\boldsymbol{f}^\mathrm{out}$ from the last layer as the output of the decoder.
This final output is upsampled by a learnable de-convolution layer and reshaped to the final triplane representation $\boldsymbol{T}$.

\subsection{Triplane-NeRF}
\label{subsec:nerf}
We employ the triplane-NeRF formulation~\citep{chan2022eg3d} and implement an $\mathrm{MLP^\mathit{nerf}}$ to predict RGB and density $\sigma$ from the point features queried from the triplane representation $\boldsymbol{T}$. 
The $\mathrm{MLP^\mathit{nerf}}$ contains multiple linear layers with ReLU~\citep{nair2010rectified} activation.
The output dimension of the $\mathrm{MLP^\mathit{nerf}}$ is $4$ where the first three dimensions are RGB colors and the last dimension corresponds to the density of the field. 
We refer to the Appendix for the details of NeRF volumetric rendering.

\subsection{Training Objectives}
\label{subsec:losses}
\ours{} produces the 3D shape from a single input image and leverages additional side views to guide the reconstruction during training. 
For each shape in the training data, we consider $(V-1)$ randomly chosen side views for supervision; we apply simple image reconstruction objectives between the $V$ rendered views $\boldsymbol{\hat{x}}$ and the ground-truth views $\boldsymbol{x}^\mathit{GT}$ (include the input view and side views). 
 More precisely, for every input image $\boldsymbol{x}$, we minimize:
\begin{align}
\mathcal{L}_\mathrm{recon}(\boldsymbol{x}) &= \frac{1}{V} {\sum_{v=1}^{V}} \left( \mathcal{L}_\mathrm{MSE}(\boldsymbol{\hat{x}}_{v},\boldsymbol{x}^\mathit{GT}_{v})+{\lambda}\mathcal{L}_\mathrm{LPIPS}(\boldsymbol{\hat{x}}_{v},\boldsymbol{x}^\mathit{GT}_{v}) \right) 
\label{eqn:loss}
\end{align}
where $\mathcal{L}_\mathrm{MSE}$ is the normalized pixel-wise L2 loss, $\mathcal{L}_\mathrm{LPIPS}$ is the perceptual image patch similarity~\citep{zhang2018lpips} and $\lambda$ is a customized weight coefficient.

\section{Experiments}
\label{sec:experiments}

\subsection{Data}
\label{subsec:data}
\ours{} relies on abundant 3D data from Objaverse~\citep{deitke2023objaverse} and MVImgNet~\citep{yu2023mvimgnet}, consisting of synthetic 3D assets and videos of objects in the real world, respectively, to learn a generalizable cross-shape 3D prior. 
For each 3D asset in Objaverse, we normalize the shape to the box $[-1,1]^3$ in world space and render $32$ random views with the same camera pointing toward the shape at arbitrary poses. The rendered images are of resolution $1024{\times}1024$, and the camera poses are sampled from a ball of radius $[1.5, 3.0]$ and with height in range $[-0.75, 1.60]$\footnote{Most of Objaverse assets have consistent $z$-axis up.}.
For each video, we utilize the extracted frames from the dataset. Since the target shape in those frames can be at random positions, we crop and resize all of them using the predicted object mask$^{\ref{rembg}}$ so that the object is at the center of the resulting frames; we adjust the camera parameters accordingly. 
Note that our method does not model background, hence we render images from Objaverse with a pure white background, and use an off-the-shelf package\footnote{Rembg package, a tool to remove image background: \href{https://pypi.org/project/rembg}{https://pypi.org/project/rembg} \label{rembg}} to remove the background of video frames. In total, we pre-processed 730,648 3D assets and 220,219 videos for training.

To evaluate the performance of \ours{} on arbitrary images, we collected novel images from Objaverse~\citep{deitke2023objaverse}, MvImgNet~\citep{yu2023mvimgnet}, ImageNet~\citep{deng2009imagenet}, Google Scanned Objects~\citep{downs2022googlescanned}, Amazon Berkeley Objects~\citep{collins2022abo}, captured new images in the real world, and generated images with Adobe Firefly\footnote{Adobe Firefly, a text-to-image generation tool: \href{https://firefly.adobe.com}{https://firefly.adobe.com}} for reconstruction. We visualize their results in Sec.~\ref{subsubsec:visualize} and Appendix.
To numerically study the design choices of our approach, we randomly acquired 50 unseen 3D shapes from the Objaverse and 50 unseen videos from the MvImgNet dataset, respectively. For each shape, we pre-process 15 reference views and pass five of them to our model one by one to reconstruct the same object, and evaluate the rendered images using all 15 reference views (see analyses in Appendix).

\subsection{Implementation Details}
\label{subsec:implement}

\paragraph{Camera Normalization}
We normalize the camera poses corresponding to the input images to facilitate the image-to-triplane modeling. Specifically, for the images rendered from synthetic 3D assets in Objaverse, regardless of the corresponding positions of the cameras, we normalize the input camera poses to position $[0,-2,0]$ with the camera vertical axis aligned with the upward $z$-axis in the world frame. For the video data, since the camera can be at an arbitrary distance from the target and the object is not at the image center, we only normalize the camera pose to $[0,-\mathit{dis},0]$ where $\mathit{dis}$ is the original distance between world origin and camera origin.

\begin{figure}[!hp]
  \centering
  \includegraphics[width=\textwidth]{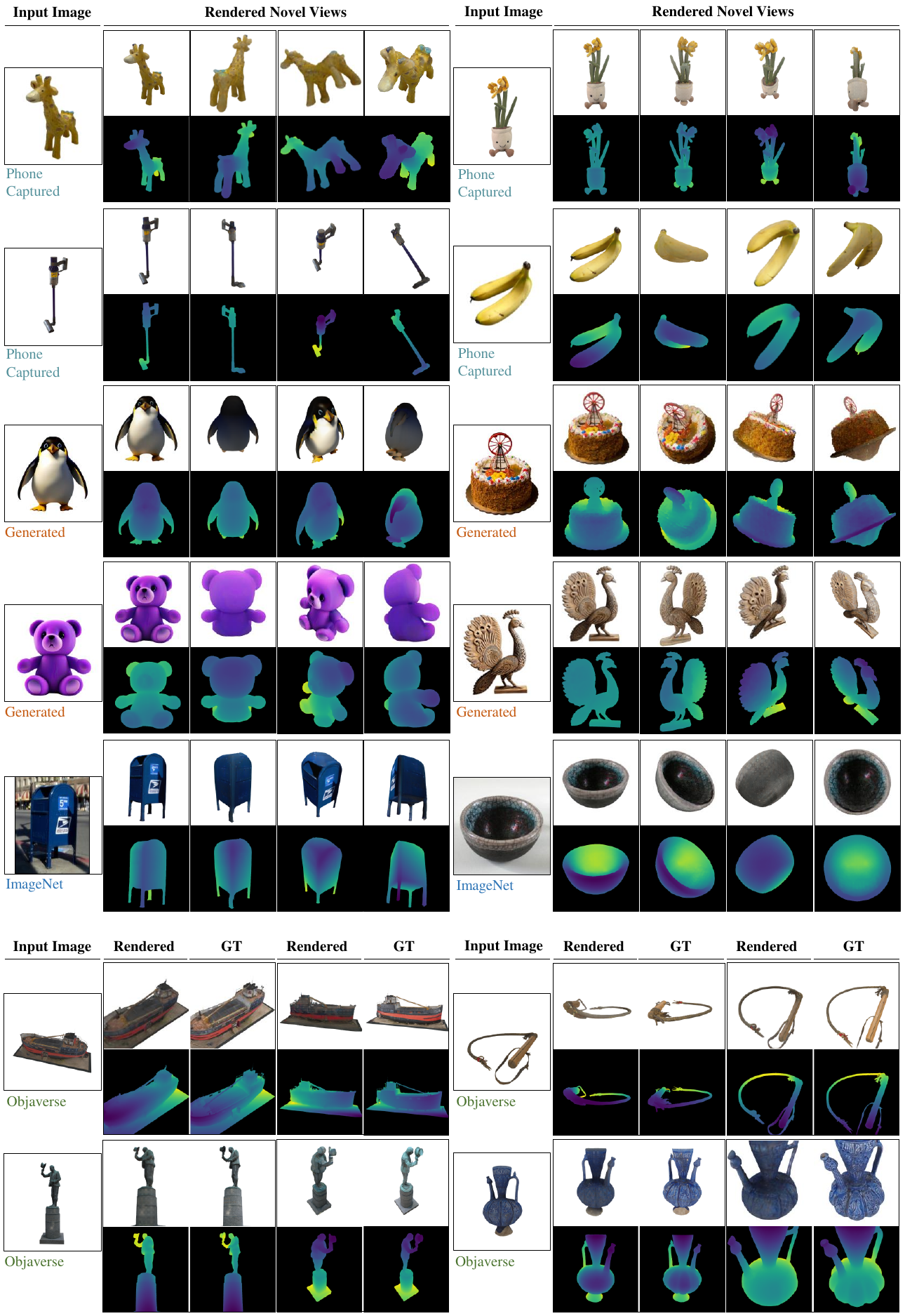}
  \vspace{-10pt}
  \caption{Rendered novel views (RGB and depth) of shapes reconstructed by our \ours{} from single images. None of the images are observed by the model during training. Generated images are created using Adobe Firefly. The last two rows compare our results to the rendered ground truth images of Objaverse objects (GT). Please zoom in for clearer visualization.
  }
  \label{fig:visualize_self}
\end{figure}

\paragraph{Network Architecture} 
We apply the ViT-B/16 model of pre-trained DINO as the image encoder, which takes $512{\times}512$ RGB images as input and produces 1025 feature tokens (1024 patch-wise features plus one \texttt{[CLS]} features) of dimension 768 ($d_E$)~\citep{caron2021dino}. 
The image-to-triplane decoder and the $\mathrm{MLP}^\mathit{nerf}$ are of 16 and 10 layers with hidden dimensions 1024 ($d_D$) and 64, respectively. 
The triplane dimension is 80 ($d_T$).
For neural rendering, LRM uniformly samples 128 points for each ray and renders $128{\times}128$ resolution images for supervision. We also use the deferred back-propagation introduced in ARF~\citep{zhang2022arf} to save GPU memory.

\paragraph{Training} 
We train \ours{} on 128 NVIDIA (40G) A100 GPUs with batch size 1024 (1024 different shapes per iteration) for 30 epochs, taking about 3 days to complete. Each epoch contains one copy of the rendered image data from Objaverse and three copies of the video frame data from MvImgNet to balance the amount of synthetic and real data. For each sample, we use $3$ randomly chosen side views (\textit{i.e.}, the total views $V=4$) to supervise the shape reconstruction, and we set the coefficient $\lambda{=}2.0$ for $\mathcal{L}_\mathrm{LPIPS}$. We apply the AdamW optimizer~\citep{loshchilov2017adamw} and set the learning rate to $4{\times}10^{-4}$ with a cosine schedule~\citep{loshchilov2016sgdr}. We numerically analyze the influence of data, training, and model hyper-parameters in the Appendix.

\paragraph{Inference} 
During inference, \ours{} takes an arbitrary image as input (squared and background removed) and assumes the unknown camera parameters to be the normalized cameras that we applied to train the Objaverse data. We query a resolution of $384{\times}384{\times}384$ points from the reconstructed triplane-NeRF and extract the mesh using Marching Cubes~\citep{lorensen1998marchingcubes}. This entire process only takes less than 5 seconds to complete on a single NVIDIA A100 GPU.

\subsection{Results}

\begin{figure}[!t]
  \centering
  \includegraphics[width=\textwidth]{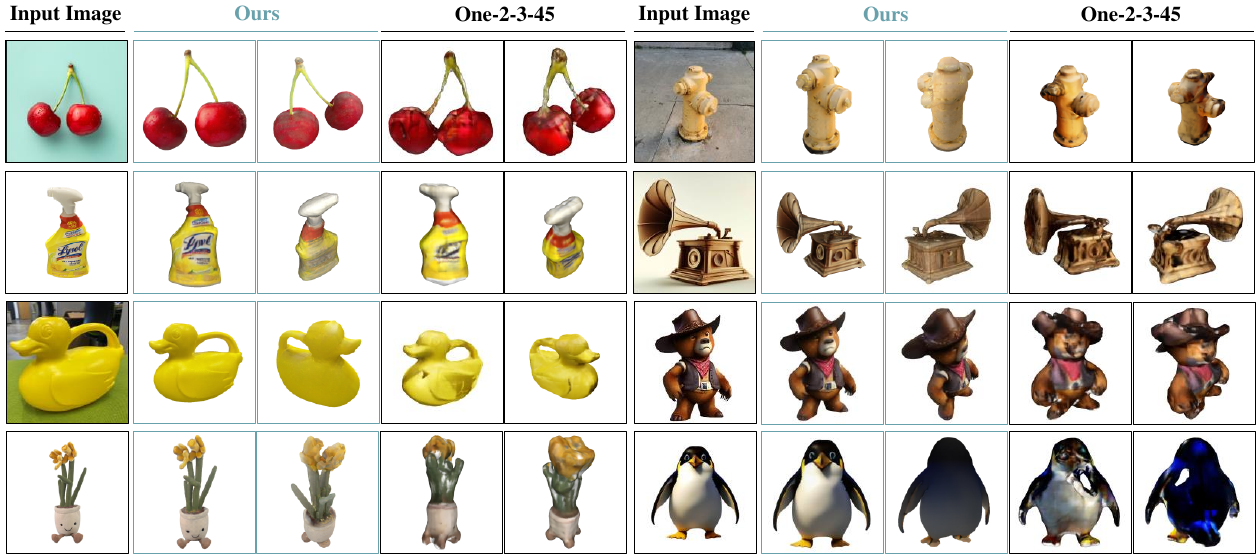}
  \vspace{-12pt}
  \caption[visualize compare]{Comparison to One-2-3-45~\citep{liu2023one2345}. To avoid cherry-picking, input images in the first three rows are selected from the examples provided in One-2-3-45's paper or demo page. None of the images are observed by our model during training. Please zoom in for clearer visualization.}
  \label{fig:visualize_compare}
\end{figure}

We visualize the novel views of shapes reconstructed from real, generated, and rendered images from various datasets (Fig.~\ref{fig:visualize_self}), compare our method with a concurrent work~\citep{liu2023one2345} (Fig.~\ref{fig:visualize_compare}), and summarize some failure cases of our method (Sec.~\ref{subsubsec:limitations}). Numerical comparisons to other methods, and analyses of data, model architecture, and supervision can be found in the Appendix.

\begin{figure}[!t]
  \centering
  \includegraphics[width=0.90\textwidth]{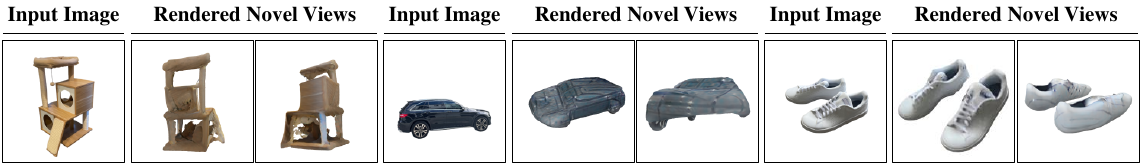}
  \vspace{-10pt}
  \caption[visualize fail]{Failure cases of our method. All three examples show blurry textures for occluded regions, and distortion due to the largely inaccurate assumption of the camera parameters.}
  \label{fig:visualize_fail}
  \vspace{-5pt}
\end{figure}

\subsubsection{Visualization}
\label{subsubsec:visualize}

Figure~\ref{fig:visualize_self} visualizes some examples of the shapes reconstructed from single images. Overall, the results show very high fidelity for diverse inputs, including real, generated, and rendered images of various subjects with distinct textures. Not only is complex geometry correctly modeled (\textit{e.g.} flower, flagon, and wipe), but also the high-frequency details, such as the texture of the wood peafowl, are preserved, both reflecting the great generalization ability of our model. From the asymmetric examples, giraffe, penguin, and bear, we can see that \ours{} can infer semantically reasonable occluded portion of the shapes, which implies effective cross-shape priors have been learned.

In Figure~\ref{fig:visualize_compare}, we compare \ours{} with One-2-3-45, a concurrent work to ours that achieves state-of-the-art single image to 3D reconstruction by generating multi-view images with 2D diffusion models~\citep{liu2023one2345}. To avoid cherry-picking, we directly test our method on the example images provided in their paper or demo page\footnote{One-2-3-45 demo page: \href{https://huggingface.co/spaces/One-2-3-45/One-2-3-45}{https://huggingface.co/spaces/One-2-3-45/One-2-3-45}.}. We can see that our method produces much sharper details and consistent surfaces. In the last row of the figure, we test One-2-3-45 with two examples used in Figure~\ref{fig:visualize_self}, showing much worse reconstruction results.

\subsubsection{Limitations}
\label{subsubsec:limitations}
Despite the high-quality single-image-to-3D results we have shown, our method still has a few limitations. First, our \ours{} tends to produce blurry textures for occluded regions, as shown in Figure ~\ref{fig:visualize_fail}. We conjecture that this is due to the fact that the single-image-to-3D problem is inherently probabilistic, i.e., multiple plausible solutions exist for the unseen region, but our model is deterministic and is likely producing averaged modes of the unseens. Second, during inference time, we assign a set of fixed camera intrinsics and extrinsics (same as our Objaverse training data) to the test images. These camera parameters may not align well with the ground truth, especially when the images are cropped and resized, causing large changes to Field-of-View (FoV) and principal points. Figure ~\ref{fig:visualize_fail} shows that incorrect assumptions of the camera parameters can lead to distorted shape reconstruction. Third, we only address images of objects without background; handling the background~\citep{zhang2020nerf++,barron2022mipnerf360}, as well as complex scenes, is beyond the scope of this work. Finally, we assume Lambertian objects and omit the view-dependent modelling~\citep{mildenhall2021nerf} in our predicted NeRF. Therefore, we cannot faithfully reconstruct the view-dependent appearance of some real-world materials, \textit{e.g.}, shiny metals, glossy ceramics, etc.

\section{Conclusion}
\label{sec:conclusion}

In this paper, we propose \ours{}, the first large transformer-based framework to learn an expressive 3D prior from a million 3D data to reconstruct objects from single images. \ours{} is very efficient in training and inference; it is a fully-differentiable network that can be trained end-to-end with simple image reconstruction losses and only takes five seconds to render a high-fidelity 3D shape, thus enabling a wide range of real-world applications. In the era of large-scale learning, we hope our idea can inspire future research to explore data-driven 3D large reconstruction models that generalize well to arbitrary in-the-wild images.

\paragraph{Future Directions} 
In addition to addressing the limitations mentioned in Sec.~\ref{subsubsec:limitations}, we suggest two future directions of our research; 
(1) Scaling up the model and training data: with the simplest transformer-based design and minimal regularization, \ours{} can be easily scaled to a larger and more capable network, including but not limited to applying a larger image encoder, adding more attention layers to the image-to-triplane decoder, and increasing the resolution of triplane representations. On the other hand, \ours{} only requires multi-view images for supervision, hence a wide range of 3D, video, and image datasets can be exploited in training. We expect both approaches to be promising in improving the model's generalization ability and the quality of reconstruction. 
(2) Extension to multimodal 3D generative models: \ours{} model builds a pathway for generating novel 3D shapes from language by leveraging a text-to-image generation model to first create 2D images. But more interestingly, we suggest the learned expressive triplane representations could be applied to directly bridge language descriptions and 3D to enable efficient text-to-3D generation and editing (\textit{e.g.}, via latent diffusion~\citep{rombach2022stablediffuse}). We will explore these ideas in our future research. 

\section*{Ethics Statement}

\ours{} proposed in this paper is a deterministic model in which, given the same image as input, the model will infer the identical 3D shape.
Unlike generative models that can be used to easily synthesize various undesirable contents (\textit{e.g.}, from language inputs), \ours{} requests the specific 2D content to exist in the first place. 
\ours{} is trained on Objaverse~\citep{deitke2023objaverse} and MvImgNet~\citep{yu2023mvimgnet} data, which mostly contain ethical content. However, given an unethical or misleading image, \ours{} could produce unethical 3D objects or 3D disinformation that may be more convincing than the 2D input images (although the reconstructed objects are less realistic than real-world objects). 

Image-to-3D reconstruction models like \ours{} hold the potential to automate tasks currently performed by 3D designers. However, it's worth noting that these tools also have the capacity to foster growth and enhance accessibility within the creative industry.

\section*{Reproducibility Statement}

Our \ours{} is built by integrating the publicly available codebases of threestudio\footnote{threestudio's GitHub page: \href{https://github.com/threestudio-project/threestudio}{https://github.com/threestudio-project/threestudio}.}~\citep{threestudio2023}, x-transformers\footnote{x-transformers's GitHub page: \href{https://github.com/lucidrains/x-transformers}{https://github.com/lucidrains/x-transformers}.}, and DINO\footnote{DINO's GitHub page: \href{https://github.com/facebookresearch/dino}{https://github.com/facebookresearch/dino}.}~\citep{caron2021dino}, and the model is trained using publicly available data from Objaverse~\citep{deitke2023objaverse} and MvImgNet~\citep{yu2023mvimgnet}.
We include very comprehensive data pre-processing, network architecture, and training details in this paper, which greatly facilitate reproducing our \ours{}.

\section*{Acknowledgment}
We want to thank Nathan Carr, Scott Cohen, Hailin Jin, Aseem Agarwala, Tong Sun for their support, and thank Duygu Ceylan, Zexiang Xu, Paul Guerrero, Chun-Hao Huang, Niloy Mitra, Radomir Mech, Vova Kim, Thibault Groueix for constructive feedback on this project. 
Hao wants to thank Xin for the inspiration as he ran on this road.
Yicong wants to thank Prof. Stephen Gould and Ms. Ziwei Wang for their great advice.

\bibliography{iclr2024_conference}

\begin{thebibliography}{104}
\providecommand{\natexlab}[1]{#1}
\providecommand{\url}[1]{\texttt{#1}}
\expandafter\ifx\csname urlstyle\endcsname\relax
  \providecommand{\doi}[1]{doi: #1}\else
  \providecommand{\doi}{doi: \begingroup \urlstyle{rm}\Url}\fi

\bibitem[Alayrac et~al.(2022)Alayrac, Donahue, Luc, Miech, Barr, Hasson, Lenc, Mensch, Millican, Reynolds, et~al.]{alayrac2022flamingo}
Jean-Baptiste Alayrac, Jeff Donahue, Pauline Luc, Antoine Miech, Iain Barr, Yana Hasson, Karel Lenc, Arthur Mensch, Katherine Millican, Malcolm Reynolds, et~al.
\newblock Flamingo: a visual language model for few-shot learning.
\newblock \emph{Advances in Neural Information Processing Systems}, 35:\penalty0 23716--23736, 2022.

\bibitem[Anciukevi{\v{c}}ius et~al.(2023)Anciukevi{\v{c}}ius, Xu, Fisher, Henderson, Bilen, Mitra, and Guerrero]{anciukevivcius2023renderdiffusion}
Titas Anciukevi{\v{c}}ius, Zexiang Xu, Matthew Fisher, Paul Henderson, Hakan Bilen, Niloy~J Mitra, and Paul Guerrero.
\newblock Renderdiffusion: Image diffusion for 3d reconstruction, inpainting and generation.
\newblock In \emph{Proceedings of the IEEE/CVF Conference on Computer Vision and Pattern Recognition}, pp.\  12608--12618, 2023.

\bibitem[Ba et~al.(2016)Ba, Kiros, and Hinton]{ba2016layernorm}
Jimmy~Lei Ba, Jamie~Ryan Kiros, and Geoffrey~E Hinton.
\newblock Layer normalization.
\newblock \emph{arXiv preprint arXiv:1607.06450}, 2016.

\bibitem[Barron et~al.(2022)Barron, Mildenhall, Verbin, Srinivasan, and Hedman]{barron2022mipnerf360}
Jonathan~T. Barron, Ben Mildenhall, Dor Verbin, Pratul~P. Srinivasan, and Peter Hedman.
\newblock Mip-nerf 360: Unbounded anti-aliased neural radiance fields.
\newblock \emph{CVPR}, 2022.

\bibitem[Bogo et~al.(2016)Bogo, Kanazawa, Lassner, Gehler, Romero, and Black]{bogo2016keepsmpl}
Federica Bogo, Angjoo Kanazawa, Christoph Lassner, Peter Gehler, Javier Romero, and Michael~J Black.
\newblock Keep it smpl: Automatic estimation of 3d human pose and shape from a single image.
\newblock In \emph{Computer Vision--ECCV 2016: 14th European Conference, Amsterdam, The Netherlands, October 11-14, 2016, Proceedings, Part V 14}, pp.\  561--578. Springer, 2016.

\bibitem[Brown et~al.(2020)Brown, Mann, Ryder, Subbiah, Kaplan, Dhariwal, Neelakantan, Shyam, Sastry, Askell, et~al.]{brown2020gpt3}
Tom Brown, Benjamin Mann, Nick Ryder, Melanie Subbiah, Jared~D Kaplan, Prafulla Dhariwal, Arvind Neelakantan, Pranav Shyam, Girish Sastry, Amanda Askell, et~al.
\newblock Language models are few-shot learners.
\newblock \emph{Advances in neural information processing systems}, 33:\penalty0 1877--1901, 2020.

\bibitem[Caron et~al.(2021)Caron, Touvron, Misra, J{\'e}gou, Mairal, Bojanowski, and Joulin]{caron2021dino}
Mathilde Caron, Hugo Touvron, Ishan Misra, Herv{\'e} J{\'e}gou, Julien Mairal, Piotr Bojanowski, and Armand Joulin.
\newblock Emerging properties in self-supervised vision transformers.
\newblock In \emph{Proceedings of the IEEE/CVF international conference on computer vision}, pp.\  9650--9660, 2021.

\bibitem[Chan et~al.(2022)Chan, Lin, Chan, Nagano, Pan, De~Mello, Gallo, Guibas, Tremblay, Khamis, et~al.]{chan2022eg3d}
Eric~R Chan, Connor~Z Lin, Matthew~A Chan, Koki Nagano, Boxiao Pan, Shalini De~Mello, Orazio Gallo, Leonidas~J Guibas, Jonathan Tremblay, Sameh Khamis, et~al.
\newblock Efficient geometry-aware 3d generative adversarial networks.
\newblock In \emph{Proceedings of the IEEE/CVF Conference on Computer Vision and Pattern Recognition}, pp.\  16123--16133, 2022.

\bibitem[Chang et~al.(2015)Chang, Funkhouser, Guibas, Hanrahan, Huang, Li, Savarese, Savva, Song, Su, et~al.]{chang2015shapenet}
Angel~X Chang, Thomas Funkhouser, Leonidas Guibas, Pat Hanrahan, Qixing Huang, Zimo Li, Silvio Savarese, Manolis Savva, Shuran Song, Hao Su, et~al.
\newblock Shapenet: An information-rich 3d model repository.
\newblock \emph{arXiv preprint arXiv:1512.03012}, 2015.

\bibitem[Chen et~al.(2022{\natexlab{a}})Chen, Xu, Geiger, Yu, and Su]{Chen2022ECCV}
Anpei Chen, Zexiang Xu, Andreas Geiger, Jingyi Yu, and Hao Su.
\newblock Tensorf: Tensorial radiance fields.
\newblock In \emph{European Conference on Computer Vision (ECCV)}, 2022{\natexlab{a}}.

\bibitem[Chen et~al.(2022{\natexlab{b}})Chen, Guo, Yi, Li, and Elhoseiny]{chen2022visualgpt}
Jun Chen, Han Guo, Kai Yi, Boyang Li, and Mohamed Elhoseiny.
\newblock Visualgpt: Data-efficient adaptation of pretrained language models for image captioning.
\newblock In \emph{Proceedings of the IEEE/CVF Conference on Computer Vision and Pattern Recognition}, pp.\  18030--18040, 2022{\natexlab{b}}.

\bibitem[Chen et~al.(2020)Chen, Li, Yu, El~Kholy, Ahmed, Gan, Cheng, and Liu]{chen2020uniter}
Yen-Chun Chen, Linjie Li, Licheng Yu, Ahmed El~Kholy, Faisal Ahmed, Zhe Gan, Yu~Cheng, and Jingjing Liu.
\newblock Uniter: Universal image-text representation learning.
\newblock In \emph{European conference on computer vision}, pp.\  104--120. Springer, 2020.

\bibitem[Chen \& Zhang(2019)Chen and Zhang]{chen2019imnet}
Zhiqin Chen and Hao Zhang.
\newblock Learning implicit fields for generative shape modeling.
\newblock In \emph{Proceedings of the IEEE/CVF Conference on Computer Vision and Pattern Recognition}, pp.\  5939--5948, 2019.

\bibitem[Chowdhery et~al.(2022)Chowdhery, Narang, Devlin, Bosma, Mishra, Roberts, Barham, Chung, Sutton, Gehrmann, et~al.]{chowdhery2022palm}
Aakanksha Chowdhery, Sharan Narang, Jacob Devlin, Maarten Bosma, Gaurav Mishra, Adam Roberts, Paul Barham, Hyung~Won Chung, Charles Sutton, Sebastian Gehrmann, et~al.
\newblock Palm: Scaling language modeling with pathways.
\newblock \emph{arXiv preprint arXiv:2204.02311}, 2022.

\bibitem[Choy et~al.(2016)Choy, Xu, Gwak, Chen, and Savarese]{choy20163dr2d2}
Christopher~B Choy, Danfei Xu, JunYoung Gwak, Kevin Chen, and Silvio Savarese.
\newblock 3d-r2n2: A unified approach for single and multi-view 3d object reconstruction.
\newblock In \emph{Computer Vision--ECCV 2016: 14th European Conference, Amsterdam, The Netherlands, October 11-14, 2016, Proceedings, Part VIII 14}, pp.\  628--644. Springer, 2016.

\bibitem[Collins et~al.(2022)Collins, Goel, Deng, Luthra, Xu, Gundogdu, Zhang, Vicente, Dideriksen, Arora, et~al.]{collins2022abo}
Jasmine Collins, Shubham Goel, Kenan Deng, Achleshwar Luthra, Leon Xu, Erhan Gundogdu, Xi~Zhang, Tomas F~Yago Vicente, Thomas Dideriksen, Himanshu Arora, et~al.
\newblock Abo: Dataset and benchmarks for real-world 3d object understanding.
\newblock In \emph{Proceedings of the IEEE/CVF Conference on Computer Vision and Pattern Recognition}, pp.\  21126--21136, 2022.

\bibitem[Dai et~al.(2017)Dai, Ruizhongtai~Qi, and Nie{\ss}ner]{dai2017shape}
Angela Dai, Charles Ruizhongtai~Qi, and Matthias Nie{\ss}ner.
\newblock Shape completion using 3d-encoder-predictor cnns and shape synthesis.
\newblock In \emph{Proceedings of the IEEE conference on computer vision and pattern recognition}, pp.\  5868--5877, 2017.

\bibitem[Deitke et~al.(2023)Deitke, Schwenk, Salvador, Weihs, Michel, VanderBilt, Schmidt, Ehsani, Kembhavi, and Farhadi]{deitke2023objaverse}
Matt Deitke, Dustin Schwenk, Jordi Salvador, Luca Weihs, Oscar Michel, Eli VanderBilt, Ludwig Schmidt, Kiana Ehsani, Aniruddha Kembhavi, and Ali Farhadi.
\newblock Objaverse: A universe of annotated 3d objects.
\newblock In \emph{Proceedings of the IEEE/CVF Conference on Computer Vision and Pattern Recognition}, pp.\  13142--13153, 2023.

\bibitem[Deng et~al.(2023)Deng, Jiang, Qi, Yan, Zhou, Guibas, Anguelov, et~al.]{deng2023nerdi}
Congyue Deng, Chiyu Jiang, Charles~R Qi, Xinchen Yan, Yin Zhou, Leonidas Guibas, Dragomir Anguelov, et~al.
\newblock Nerdi: Single-view nerf synthesis with language-guided diffusion as general image priors.
\newblock In \emph{Proceedings of the IEEE/CVF Conference on Computer Vision and Pattern Recognition}, pp.\  20637--20647, 2023.

\bibitem[Deng et~al.(2009)Deng, Dong, Socher, Li, Li, and Fei-Fei]{deng2009imagenet}
Jia Deng, Wei Dong, Richard Socher, Li-Jia Li, Kai Li, and Li~Fei-Fei.
\newblock Imagenet: A large-scale hierarchical image database.
\newblock In \emph{2009 IEEE conference on computer vision and pattern recognition}, pp.\  248--255. Ieee, 2009.

\bibitem[Devlin et~al.(2018)Devlin, Chang, Lee, and Toutanova]{devlin2018bert}
Jacob Devlin, Ming-Wei Chang, Kenton Lee, and Kristina Toutanova.
\newblock Bert: Pre-training of deep bidirectional transformers for language understanding.
\newblock \emph{arXiv preprint arXiv:1810.04805}, 2018.

\bibitem[Dosovitskiy et~al.(2020)Dosovitskiy, Beyer, Kolesnikov, Weissenborn, Zhai, Unterthiner, Dehghani, Minderer, Heigold, Gelly, et~al.]{dosovitskiy2020vit}
Alexey Dosovitskiy, Lucas Beyer, Alexander Kolesnikov, Dirk Weissenborn, Xiaohua Zhai, Thomas Unterthiner, Mostafa Dehghani, Matthias Minderer, Georg Heigold, Sylvain Gelly, et~al.
\newblock An image is worth 16x16 words: Transformers for image recognition at scale.
\newblock \emph{arXiv preprint arXiv:2010.11929}, 2020.

\bibitem[Downs et~al.(2022)Downs, Francis, Koenig, Kinman, Hickman, Reymann, McHugh, and Vanhoucke]{downs2022googlescanned}
Laura Downs, Anthony Francis, Nate Koenig, Brandon Kinman, Ryan Hickman, Krista Reymann, Thomas~B McHugh, and Vincent Vanhoucke.
\newblock Google scanned objects: A high-quality dataset of 3d scanned household items.
\newblock In \emph{2022 International Conference on Robotics and Automation (ICRA)}, pp.\  2553--2560. IEEE, 2022.

\bibitem[Fan et~al.(2017)Fan, Su, and Guibas]{fan2017pointset}
Haoqiang Fan, Hao Su, and Leonidas~J Guibas.
\newblock A point set generation network for 3d object reconstruction from a single image.
\newblock In \emph{Proceedings of the IEEE conference on computer vision and pattern recognition}, pp.\  605--613, 2017.

\bibitem[Gao et~al.(2022)Gao, Shen, Wang, Chen, Yin, Li, Litany, Gojcic, and Fidler]{gao2022get3d}
Jun Gao, Tianchang Shen, Zian Wang, Wenzheng Chen, Kangxue Yin, Daiqing Li, Or~Litany, Zan Gojcic, and Sanja Fidler.
\newblock Get3d: A generative model of high quality 3d textured shapes learned from images.
\newblock \emph{Advances In Neural Information Processing Systems}, 35:\penalty0 31841--31854, 2022.

\bibitem[Girdhar et~al.(2016)Girdhar, Fouhey, Rodriguez, and Gupta]{girdhar2016learning}
Rohit Girdhar, David~F Fouhey, Mikel Rodriguez, and Abhinav Gupta.
\newblock Learning a predictable and generative vector representation for objects.
\newblock In \emph{Computer Vision--ECCV 2016: 14th European Conference, Amsterdam, The Netherlands, October 11-14, 2016, Proceedings, Part VI 14}, pp.\  484--499. Springer, 2016.

\bibitem[Girdhar et~al.(2023)Girdhar, El-Nouby, Liu, Singh, Alwala, Joulin, and Misra]{girdhar2023imagebind}
Rohit Girdhar, Alaaeldin El-Nouby, Zhuang Liu, Mannat Singh, Kalyan~Vasudev Alwala, Armand Joulin, and Ishan Misra.
\newblock Imagebind: One embedding space to bind them all.
\newblock In \emph{Proceedings of the IEEE/CVF Conference on Computer Vision and Pattern Recognition}, pp.\  15180--15190, 2023.

\bibitem[Gkioxari et~al.(2019)Gkioxari, Malik, and Johnson]{gkioxari2019meshrcnn}
Georgia Gkioxari, Jitendra Malik, and Justin Johnson.
\newblock Mesh r-cnn.
\newblock In \emph{Proceedings of the IEEE/CVF international conference on computer vision}, pp.\  9785--9795, 2019.

\bibitem[Goel et~al.(2020)Goel, Kanazawa, and Malik]{goel2020without}
Shubham Goel, Angjoo Kanazawa, and Jitendra Malik.
\newblock Shape and viewpoint without keypoints.
\newblock In \emph{Computer Vision--ECCV 2020: 16th European Conference, Glasgow, UK, August 23--28, 2020, Proceedings, Part XV 16}, pp.\  88--104. Springer, 2020.

\bibitem[Goel et~al.(2023)Goel, Pavlakos, Rajasegaran, Kanazawa, and Malik]{goel2023humans4d}
Shubham Goel, Georgios Pavlakos, Jathushan Rajasegaran, Angjoo Kanazawa, and Jitendra Malik.
\newblock Humans in 4d: Reconstructing and tracking humans with transformers.
\newblock In \emph{Proceedings of the IEEE/CVF International Conference on Computer Vision}, pp.\  14783--14794, 2023.

\bibitem[Guo et~al.(2023)Guo, Liu, Shao, Laforte, Voleti, Luo, Chen, Zou, Wang, Cao, and Zhang]{threestudio2023}
Yuan-Chen Guo, Ying-Tian Liu, Ruizhi Shao, Christian Laforte, Vikram Voleti, Guan Luo, Chia-Hao Chen, Zi-Xin Zou, Chen Wang, Yan-Pei Cao, and Song-Hai Zhang.
\newblock threestudio: A unified framework for 3d content generation.
\newblock \url{https://github.com/threestudio-project/threestudio}, 2023.

\bibitem[He et~al.(2016)He, Zhang, Ren, and Sun]{he2016resnet}
Kaiming He, Xiangyu Zhang, Shaoqing Ren, and Jian Sun.
\newblock Deep residual learning for image recognition.
\newblock In \emph{Proceedings of the IEEE conference on computer vision and pattern recognition}, pp.\  770--778, 2016.

\bibitem[Hendrycks \& Gimpel(2023)Hendrycks and Gimpel]{hendrycks2023gelu}
Dan Hendrycks and Kevin Gimpel.
\newblock Gaussian error linear units (gelus), 2023.

\bibitem[Hong et~al.(2023)Hong, Zhen, Chen, Zheng, Du, Chen, and Gan]{hong20233dllm}
Yining Hong, Haoyu Zhen, Peihao Chen, Shuhong Zheng, Yilun Du, Zhenfang Chen, and Chuang Gan.
\newblock 3d-llm: Injecting the 3d world into large language models.
\newblock \emph{arXiv}, 2023.

\bibitem[Huang \& Belongie(2017)Huang and Belongie]{huang2017arbitrary}
Xun Huang and Serge Belongie.
\newblock Arbitrary style transfer in real-time with adaptive instance normalization.
\newblock In \emph{Proceedings of the IEEE international conference on computer vision}, pp.\  1501--1510, 2017.

\bibitem[Jaegle et~al.(2021)Jaegle, Gimeno, Brock, Vinyals, Zisserman, and Carreira]{jaegle2021perceiver}
Andrew Jaegle, Felix Gimeno, Andy Brock, Oriol Vinyals, Andrew Zisserman, and Joao Carreira.
\newblock Perceiver: General perception with iterative attention.
\newblock In \emph{International conference on machine learning}, pp.\  4651--4664. PMLR, 2021.

\bibitem[Jang \& Agapito(2021)Jang and Agapito]{jang2021codenerf}
Wonbong Jang and Lourdes Agapito.
\newblock Codenerf: Disentangled neural radiance fields for object categories.
\newblock In \emph{Proceedings of the IEEE/CVF International Conference on Computer Vision}, pp.\  12949--12958, 2021.

\bibitem[Jun \& Nichol(2023)Jun and Nichol]{jun2023shape}
Heewoo Jun and Alex Nichol.
\newblock Shap-e: Generating conditional 3d implicit functions.
\newblock \emph{arXiv preprint arXiv:2305.02463}, 2023.

\bibitem[Kanazawa et~al.(2018)Kanazawa, Tulsiani, Efros, and Malik]{kanazawa2018learning}
Angjoo Kanazawa, Shubham Tulsiani, Alexei~A Efros, and Jitendra Malik.
\newblock Learning category-specific mesh reconstruction from image collections.
\newblock In \emph{Proceedings of the European Conference on Computer Vision (ECCV)}, pp.\  371--386, 2018.

\bibitem[Kerr et~al.(2023)Kerr, Kim, Goldberg, Kanazawa, and Tancik]{kerr2023lerf}
Justin Kerr, Chung~Min Kim, Ken Goldberg, Angjoo Kanazawa, and Matthew Tancik.
\newblock Lerf: Language embedded radiance fields.
\newblock \emph{arXiv preprint arXiv:2303.09553}, 2023.

\bibitem[Kulkarni et~al.(2020)Kulkarni, Gupta, Fouhey, and Tulsiani]{kulkarni2020articulation}
Nilesh Kulkarni, Abhinav Gupta, David~F Fouhey, and Shubham Tulsiani.
\newblock Articulation-aware canonical surface mapping.
\newblock In \emph{Proceedings of the IEEE/CVF Conference on Computer Vision and Pattern Recognition}, pp.\  452--461, 2020.

\bibitem[Li et~al.(2023{\natexlab{a}})Li, Tan, Zhang, Xu, Luan, Xu, Hong, Sunkavalli, Shakhnarovich, and Bi]{li2023instant3d}
Jiahao Li, Hao Tan, Kai Zhang, Zexiang Xu, Fujun Luan, Yinghao Xu, Yicong Hong, Kalyan Sunkavalli, Greg Shakhnarovich, and Sai Bi.
\newblock Instant3d: Fast text-to-3d with sparse-view generation and large reconstruction model.
\newblock \emph{arXiv preprint arXiv:2311.06214}, 2023{\natexlab{a}}.

\bibitem[Li et~al.(2022)Li, Li, Xiong, and Hoi]{li2022blip}
Junnan Li, Dongxu Li, Caiming Xiong, and Steven Hoi.
\newblock Blip: Bootstrapping language-image pre-training for unified vision-language understanding and generation.
\newblock In \emph{International Conference on Machine Learning}, pp.\  12888--12900. PMLR, 2022.

\bibitem[Li et~al.(2023{\natexlab{b}})Li, Li, Savarese, and Hoi]{li2023blip2}
Junnan Li, Dongxu Li, Silvio Savarese, and Steven Hoi.
\newblock Blip-2: Bootstrapping language-image pre-training with frozen image encoders and large language models.
\newblock \emph{arXiv preprint arXiv:2301.12597}, 2023{\natexlab{b}}.

\bibitem[Li et~al.(2020)Li, Liu, Kim, De~Mello, Jampani, Yang, and Kautz]{li2020self}
Xueting Li, Sifei Liu, Kihwan Kim, Shalini De~Mello, Varun Jampani, Ming-Hsuan Yang, and Jan Kautz.
\newblock Self-supervised single-view 3d reconstruction via semantic consistency.
\newblock In \emph{Computer Vision--ECCV 2020: 16th European Conference, Glasgow, UK, August 23--28, 2020, Proceedings, Part XIV 16}, pp.\  677--693. Springer, 2020.

\bibitem[Liu et~al.(2023{\natexlab{a}})Liu, Xu, Jin, Chen, Xu, Su, et~al.]{liu2023one2345}
Minghua Liu, Chao Xu, Haian Jin, Linghao Chen, Zexiang Xu, Hao Su, et~al.
\newblock One-2-3-45: Any single image to 3d mesh in 45 seconds without per-shape optimization.
\newblock \emph{arXiv preprint arXiv:2306.16928}, 2023{\natexlab{a}}.

\bibitem[Liu et~al.(2023{\natexlab{b}})Liu, Wu, Van~Hoorick, Tokmakov, Zakharov, and Vondrick]{liu2023zero123}
Ruoshi Liu, Rundi Wu, Basile Van~Hoorick, Pavel Tokmakov, Sergey Zakharov, and Carl Vondrick.
\newblock Zero-1-to-3: Zero-shot one image to 3d object.
\newblock \emph{arXiv preprint arXiv:2303.11328}, 2023{\natexlab{b}}.

\bibitem[Lorensen \& Cline(1998)Lorensen and Cline]{lorensen1998marchingcubes}
William~E Lorensen and Harvey~E Cline.
\newblock Marching cubes: A high resolution 3d surface construction algorithm.
\newblock In \emph{Seminal graphics: pioneering efforts that shaped the field}, pp.\  347--353. 1998.

\bibitem[Loshchilov \& Hutter(2016)Loshchilov and Hutter]{loshchilov2016sgdr}
Ilya Loshchilov and Frank Hutter.
\newblock Sgdr: Stochastic gradient descent with warm restarts.
\newblock \emph{arXiv preprint arXiv:1608.03983}, 2016.

\bibitem[Loshchilov \& Hutter(2017)Loshchilov and Hutter]{loshchilov2017adamw}
Ilya Loshchilov and Frank Hutter.
\newblock Decoupled weight decay regularization.
\newblock \emph{arXiv preprint arXiv:1711.05101}, 2017.

\bibitem[Luo et~al.(2023)Luo, Rockwell, Lee, and Johnson]{luo2023cap3d}
Tiange Luo, Chris Rockwell, Honglak Lee, and Justin Johnson.
\newblock Scalable 3d captioning with pretrained models.
\newblock \emph{arXiv preprint arXiv:2306.07279}, 2023.

\bibitem[Mandikal et~al.(2018)Mandikal, Navaneet, Agarwal, and Babu]{mandikal20183dlmnet}
Priyanka Mandikal, KL~Navaneet, Mayank Agarwal, and R~Venkatesh Babu.
\newblock 3d-lmnet: Latent embedding matching for accurate and diverse 3d point cloud reconstruction from a single image.
\newblock \emph{arXiv preprint arXiv:1807.07796}, 2018.

\bibitem[Melas-Kyriazi et~al.(2023)Melas-Kyriazi, Laina, Rupprecht, and Vedaldi]{melas2023realfusion}
Luke Melas-Kyriazi, Iro Laina, Christian Rupprecht, and Andrea Vedaldi.
\newblock Realfusion: 360deg reconstruction of any object from a single image.
\newblock In \emph{Proceedings of the IEEE/CVF Conference on Computer Vision and Pattern Recognition}, pp.\  8446--8455, 2023.

\bibitem[Mescheder et~al.(2019)Mescheder, Oechsle, Niemeyer, Nowozin, and Geiger]{mescheder2019occupancy}
Lars Mescheder, Michael Oechsle, Michael Niemeyer, Sebastian Nowozin, and Andreas Geiger.
\newblock Occupancy networks: Learning 3d reconstruction in function space.
\newblock In \emph{Proceedings of the IEEE/CVF conference on computer vision and pattern recognition}, pp.\  4460--4470, 2019.

\bibitem[Metzer et~al.(2023)Metzer, Richardson, Patashnik, Giryes, and Cohen-Or]{metzer2023latentnerf}
Gal Metzer, Elad Richardson, Or~Patashnik, Raja Giryes, and Daniel Cohen-Or.
\newblock Latent-nerf for shape-guided generation of 3d shapes and textures.
\newblock In \emph{Proceedings of the IEEE/CVF Conference on Computer Vision and Pattern Recognition}, pp.\  12663--12673, 2023.

\bibitem[Mildenhall et~al.(2021)Mildenhall, Srinivasan, Tancik, Barron, Ramamoorthi, and Ng]{mildenhall2021nerf}
Ben Mildenhall, Pratul~P Srinivasan, Matthew Tancik, Jonathan~T Barron, Ravi Ramamoorthi, and Ren Ng.
\newblock Nerf: Representing scenes as neural radiance fields for view synthesis.
\newblock \emph{Communications of the ACM}, 65\penalty0 (1):\penalty0 99--106, 2021.

\bibitem[Mittal et~al.(2022)Mittal, Cheng, Singh, and Tulsiani]{mittal2022autosdf}
Paritosh Mittal, Yen-Chi Cheng, Maneesh Singh, and Shubham Tulsiani.
\newblock Autosdf: Shape priors for 3d completion, reconstruction and generation.
\newblock In \emph{Proceedings of the IEEE/CVF Conference on Computer Vision and Pattern Recognition}, pp.\  306--315, 2022.

\bibitem[M{\"u}ller et~al.(2022)M{\"u}ller, Simonelli, Porzi, Bulo, Nie{\ss}ner, and Kontschieder]{muller2022autorf}
Norman M{\"u}ller, Andrea Simonelli, Lorenzo Porzi, Samuel~Rota Bulo, Matthias Nie{\ss}ner, and Peter Kontschieder.
\newblock Autorf: Learning 3d object radiance fields from single view observations.
\newblock In \emph{Proceedings of the IEEE/CVF Conference on Computer Vision and Pattern Recognition}, pp.\  3971--3980, 2022.

\bibitem[M\"uller et~al.(2022)M\"uller, Evans, Schied, and Keller]{mueller2022instant}
Thomas M\"uller, Alex Evans, Christoph Schied, and Alexander Keller.
\newblock Instant neural graphics primitives with a multiresolution hash encoding.
\newblock \emph{ACM Trans. Graph.}, 41\penalty0 (4):\penalty0 102:1--102:15, July 2022.
\newblock \doi{10.1145/3528223.3530127}.
\newblock URL \url{https://doi.org/10.1145/3528223.3530127}.

\bibitem[Nair \& Hinton(2010)Nair and Hinton]{nair2010rectified}
Vinod Nair and Geoffrey~E Hinton.
\newblock Rectified linear units improve restricted boltzmann machines.
\newblock In \emph{Proceedings of the 27th international conference on machine learning (ICML-10)}, pp.\  807--814, 2010.

\bibitem[Nichol et~al.(2022)Nichol, Jun, Dhariwal, Mishkin, and Chen]{nichol2022pointe}
Alex Nichol, Heewoo Jun, Prafulla Dhariwal, Pamela Mishkin, and Mark Chen.
\newblock Point-e: A system for generating 3d point clouds from complex prompts.
\newblock \emph{arXiv preprint arXiv:2212.08751}, 2022.

\bibitem[Niemeyer et~al.(2020)Niemeyer, Mescheder, Oechsle, and Geiger]{niemeyer2020differentiable}
Michael Niemeyer, Lars Mescheder, Michael Oechsle, and Andreas Geiger.
\newblock Differentiable volumetric rendering: Learning implicit 3d representations without 3d supervision.
\newblock In \emph{Proceedings of the IEEE/CVF Conference on Computer Vision and Pattern Recognition}, pp.\  3504--3515, 2020.

\bibitem[Novotny et~al.(2019)Novotny, Ravi, Graham, Neverova, and Vedaldi]{novotny2019c3dpo}
David Novotny, Nikhila Ravi, Benjamin Graham, Natalia Neverova, and Andrea Vedaldi.
\newblock C3dpo: Canonical 3d pose networks for non-rigid structure from motion.
\newblock In \emph{Proceedings of the IEEE/CVF International Conference on Computer Vision}, pp.\  7688--7697, 2019.

\bibitem[OpenAI(2023)]{openai2023gpt4}
OpenAI.
\newblock Gpt-4 technical report, 2023.

\bibitem[Park et~al.(2019)Park, Florence, Straub, Newcombe, and Lovegrove]{park2019deepsdf}
Jeong~Joon Park, Peter Florence, Julian Straub, Richard Newcombe, and Steven Lovegrove.
\newblock Deepsdf: Learning continuous signed distance functions for shape representation.
\newblock In \emph{Proceedings of the IEEE/CVF conference on computer vision and pattern recognition}, pp.\  165--174, 2019.

\bibitem[Paszke et~al.(2019)Paszke, Gross, Massa, Lerer, Bradbury, Chanan, Killeen, Lin, Gimelshein, Antiga, et~al.]{paszke2019pytorch}
Adam Paszke, Sam Gross, Francisco Massa, Adam Lerer, James Bradbury, Gregory Chanan, Trevor Killeen, Zeming Lin, Natalia Gimelshein, Luca Antiga, et~al.
\newblock Pytorch: An imperative style, high-performance deep learning library.
\newblock \emph{Advances in neural information processing systems}, 32, 2019.

\bibitem[Peebles \& Xie(2022)Peebles and Xie]{peebles2022dit}
William Peebles and Saining Xie.
\newblock Scalable diffusion models with transformers.
\newblock \emph{arXiv preprint arXiv:2212.09748}, 2022.

\bibitem[Poole et~al.(2022)Poole, Jain, Barron, and Mildenhall]{poole2022dreamfusion}
Ben Poole, Ajay Jain, Jonathan~T Barron, and Ben Mildenhall.
\newblock Dreamfusion: Text-to-3d using 2d diffusion.
\newblock \emph{arXiv preprint arXiv:2209.14988}, 2022.

\bibitem[Qian et~al.(2023)Qian, Mai, Hamdi, Ren, Siarohin, Li, Lee, Skorokhodov, Wonka, Tulyakov, et~al.]{qian2023magic123}
Guocheng Qian, Jinjie Mai, Abdullah Hamdi, Jian Ren, Aliaksandr Siarohin, Bing Li, Hsin-Ying Lee, Ivan Skorokhodov, Peter Wonka, Sergey Tulyakov, et~al.
\newblock Magic123: One image to high-quality 3d object generation using both 2d and 3d diffusion priors.
\newblock \emph{arXiv preprint arXiv:2306.17843}, 2023.

\bibitem[Radford et~al.(2019)Radford, Wu, Child, Luan, Amodei, Sutskever, et~al.]{radford2019language}
Alec Radford, Jeffrey Wu, Rewon Child, David Luan, Dario Amodei, Ilya Sutskever, et~al.
\newblock Language models are unsupervised multitask learners.
\newblock \emph{OpenAI blog}, 1\penalty0 (8):\penalty0 9, 2019.

\bibitem[Radford et~al.(2021)Radford, Kim, Hallacy, Ramesh, Goh, Agarwal, Sastry, Askell, Mishkin, Clark, et~al.]{radford2021clip}
Alec Radford, Jong~Wook Kim, Chris Hallacy, Aditya Ramesh, Gabriel Goh, Sandhini Agarwal, Girish Sastry, Amanda Askell, Pamela Mishkin, Jack Clark, et~al.
\newblock Learning transferable visual models from natural language supervision.
\newblock In \emph{International conference on machine learning}, pp.\  8748--8763. PMLR, 2021.

\bibitem[Ramesh et~al.(2021)Ramesh, Pavlov, Goh, Gray, Voss, Radford, Chen, and Sutskever]{ramesh2021dalle}
Aditya Ramesh, Mikhail Pavlov, Gabriel Goh, Scott Gray, Chelsea Voss, Alec Radford, Mark Chen, and Ilya Sutskever.
\newblock Zero-shot text-to-image generation.
\newblock In \emph{International Conference on Machine Learning}, pp.\  8821--8831. PMLR, 2021.

\bibitem[Ramesh et~al.(2022)Ramesh, Dhariwal, Nichol, Chu, and Chen]{ramesh2022dalle2}
Aditya Ramesh, Prafulla Dhariwal, Alex Nichol, Casey Chu, and Mark Chen.
\newblock Hierarchical text-conditional image generation with clip latents.
\newblock \emph{arXiv preprint arXiv:2204.06125}, 2022.

\bibitem[Reizenstein et~al.(2021)Reizenstein, Shapovalov, Henzler, Sbordone, Labatut, and Novotny]{reizenstein2021co3d}
Jeremy Reizenstein, Roman Shapovalov, Philipp Henzler, Luca Sbordone, Patrick Labatut, and David Novotny.
\newblock Common objects in 3d: Large-scale learning and evaluation of real-life 3d category reconstruction.
\newblock In \emph{Proceedings of the IEEE/CVF International Conference on Computer Vision}, pp.\  10901--10911, 2021.

\bibitem[Rombach et~al.(2022)Rombach, Blattmann, Lorenz, Esser, and Ommer]{rombach2022stablediffuse}
Robin Rombach, Andreas Blattmann, Dominik Lorenz, Patrick Esser, and Bj{\"o}rn Ommer.
\newblock High-resolution image synthesis with latent diffusion models.
\newblock In \emph{Proceedings of the IEEE/CVF conference on computer vision and pattern recognition}, pp.\  10684--10695, 2022.

\bibitem[Roth et~al.(2016)Roth, Tong, and Liu]{roth2016adaptive}
Joseph Roth, Yiying Tong, and Xiaoming Liu.
\newblock Adaptive 3d face reconstruction from unconstrained photo collections.
\newblock In \emph{Proceedings of the IEEE conference on computer vision and pattern recognition}, pp.\  4197--4206, 2016.

\bibitem[Saharia et~al.(2022)Saharia, Chan, Saxena, Li, Whang, Denton, Ghasemipour, Gontijo~Lopes, Karagol~Ayan, Salimans, et~al.]{saharia2022imagen}
Chitwan Saharia, William Chan, Saurabh Saxena, Lala Li, Jay Whang, Emily~L Denton, Kamyar Ghasemipour, Raphael Gontijo~Lopes, Burcu Karagol~Ayan, Tim Salimans, et~al.
\newblock Photorealistic text-to-image diffusion models with deep language understanding.
\newblock \emph{Advances in Neural Information Processing Systems}, 35:\penalty0 36479--36494, 2022.

\bibitem[Sajjadi et~al.(2022)Sajjadi, Meyer, Pot, Bergmann, Greff, Radwan, Vora, Lu{\v{c}}i{\'c}, Duckworth, Dosovitskiy, et~al.]{sajjadi2022srt}
Mehdi~SM Sajjadi, Henning Meyer, Etienne Pot, Urs Bergmann, Klaus Greff, Noha Radwan, Suhani Vora, Mario Lu{\v{c}}i{\'c}, Daniel Duckworth, Alexey Dosovitskiy, et~al.
\newblock Scene representation transformer: Geometry-free novel view synthesis through set-latent scene representations.
\newblock In \emph{Proceedings of the IEEE/CVF Conference on Computer Vision and Pattern Recognition}, pp.\  6229--6238, 2022.

\bibitem[Shen et~al.(2023{\natexlab{a}})Shen, Yan, Qi, Najibi, Deng, Guibas, Zhou, and Anguelov]{shen2023gina3d}
Bokui Shen, Xinchen Yan, Charles~R Qi, Mahyar Najibi, Boyang Deng, Leonidas Guibas, Yin Zhou, and Dragomir Anguelov.
\newblock Gina-3d: Learning to generate implicit neural assets in the wild.
\newblock In \emph{Proceedings of the IEEE/CVF Conference on Computer Vision and Pattern Recognition}, pp.\  4913--4926, 2023{\natexlab{a}}.

\bibitem[Shen et~al.(2023{\natexlab{b}})Shen, Yang, and Wang]{shen2023anything3d}
Qiuhong Shen, Xingyi Yang, and Xinchao Wang.
\newblock Anything-3d: Towards single-view anything reconstruction in the wild.
\newblock \emph{arXiv preprint arXiv:2304.10261}, 2023{\natexlab{b}}.

\bibitem[Singh et~al.(2022)Singh, Hu, Goswami, Couairon, Galuba, Rohrbach, and Kiela]{singh2022flava}
Amanpreet Singh, Ronghang Hu, Vedanuj Goswami, Guillaume Couairon, Wojciech Galuba, Marcus Rohrbach, and Douwe Kiela.
\newblock Flava: A foundational language and vision alignment model.
\newblock In \emph{Proceedings of the IEEE/CVF Conference on Computer Vision and Pattern Recognition}, pp.\  15638--15650, 2022.

\bibitem[Sun et~al.(2022)Sun, Sun, and Chen]{SunSC22}
Cheng Sun, Min Sun, and Hwann{-}Tzong Chen.
\newblock Direct voxel grid optimization: Super-fast convergence for radiance fields reconstruction.
\newblock In \emph{CVPR}, 2022.

\bibitem[Tan \& Bansal(2019)Tan and Bansal]{tan2019lxmert}
Hao Tan and Mohit Bansal.
\newblock Lxmert: Learning cross-modality encoder representations from transformers.
\newblock \emph{arXiv preprint arXiv:1908.07490}, 2019.

\bibitem[Tang et~al.(2023)Tang, Wang, Zhang, Zhang, Yi, Ma, and Chen]{tang2023makeit3d}
Junshu Tang, Tengfei Wang, Bo~Zhang, Ting Zhang, Ran Yi, Lizhuang Ma, and Dong Chen.
\newblock Make-it-3d: High-fidelity 3d creation from a single image with diffusion prior.
\newblock \emph{arXiv preprint arXiv:2303.14184}, 2023.

\bibitem[Touvron et~al.(2023)Touvron, Lavril, Izacard, Martinet, Lachaux, Lacroix, Rozi{\`e}re, Goyal, Hambro, Azhar, et~al.]{touvron2023llama}
Hugo Touvron, Thibaut Lavril, Gautier Izacard, Xavier Martinet, Marie-Anne Lachaux, Timoth{\'e}e Lacroix, Baptiste Rozi{\`e}re, Naman Goyal, Eric Hambro, Faisal Azhar, et~al.
\newblock Llama: Open and efficient foundation language models.
\newblock \emph{arXiv preprint arXiv:2302.13971}, 2023.

\bibitem[Tulsiani et~al.(2017)Tulsiani, Zhou, Efros, and Malik]{tulsiani2017drc}
Shubham Tulsiani, Tinghui Zhou, Alexei~A Efros, and Jitendra Malik.
\newblock Multi-view supervision for single-view reconstruction via differentiable ray consistency.
\newblock In \emph{Proceedings of the IEEE conference on computer vision and pattern recognition}, pp.\  2626--2634, 2017.

\bibitem[Vaswani et~al.(2017)Vaswani, Shazeer, Parmar, Uszkoreit, Jones, Gomez, Kaiser, and Polosukhin]{vaswani2017attention}
Ashish Vaswani, Noam Shazeer, Niki Parmar, Jakob Uszkoreit, Llion Jones, Aidan~N Gomez, {\L}ukasz Kaiser, and Illia Polosukhin.
\newblock Attention is all you need.
\newblock \emph{Advances in neural information processing systems}, 30, 2017.

\bibitem[Wang et~al.(2018)Wang, Zhang, Li, Fu, Liu, and Jiang]{wang2018pixel2mesh}
Nanyang Wang, Yinda Zhang, Zhuwen Li, Yanwei Fu, Wei Liu, and Yu-Gang Jiang.
\newblock Pixel2mesh: Generating 3d mesh models from single rgb images.
\newblock In \emph{Proceedings of the European conference on computer vision (ECCV)}, pp.\  52--67, 2018.

\bibitem[Wang et~al.(2022)Wang, Li, Li, He, Huang, Zhao, Zhang, Xu, Liu, Wang, et~al.]{wang2022internvideo}
Yi~Wang, Kunchang Li, Yizhuo Li, Yinan He, Bingkun Huang, Zhiyu Zhao, Hongjie Zhang, Jilan Xu, Yi~Liu, Zun Wang, et~al.
\newblock Internvideo: General video foundation models via generative and discriminative learning.
\newblock \emph{arXiv preprint arXiv:2212.03191}, 2022.

\bibitem[Wang et~al.(2004)Wang, Bovik, Sheikh, and Simoncelli]{wang2004ssim}
Zhou Wang, Alan~C Bovik, Hamid~R Sheikh, and Eero~P Simoncelli.
\newblock Image quality assessment: from error visibility to structural similarity.
\newblock \emph{IEEE transactions on image processing}, 13\penalty0 (4):\penalty0 600--612, 2004.

\bibitem[Wu et~al.(2023)Wu, Johnson, Malik, Feichtenhofer, and Gkioxari]{wu2023mcc}
Chao-Yuan Wu, Justin Johnson, Jitendra Malik, Christoph Feichtenhofer, and Georgia Gkioxari.
\newblock Multiview compressive coding for 3d reconstruction.
\newblock In \emph{Proceedings of the IEEE/CVF Conference on Computer Vision and Pattern Recognition}, pp.\  9065--9075, 2023.

\bibitem[Wu et~al.(2020)Wu, Zhuang, Xu, Zhang, and Chen]{wu2020pq}
Rundi Wu, Yixin Zhuang, Kai Xu, Hao Zhang, and Baoquan Chen.
\newblock Pq-net: A generative part seq2seq network for 3d shapes.
\newblock In \emph{Proceedings of the IEEE/CVF Conference on Computer Vision and Pattern Recognition}, pp.\  829--838, 2020.

\bibitem[Xu et~al.(2023)Xu, Jiang, Wang, Fan, Wang, and Wang]{xu2023neurallift360}
Dejia Xu, Yifan Jiang, Peihao Wang, Zhiwen Fan, Yi~Wang, and Zhangyang Wang.
\newblock Neurallift-360: Lifting an in-the-wild 2d photo to a 3d object with 360deg views.
\newblock In \emph{Proceedings of the IEEE/CVF Conference on Computer Vision and Pattern Recognition}, pp.\  4479--4489, 2023.

\bibitem[Xu et~al.(2019)Xu, Wang, Ceylan, Mech, and Neumann]{xu2019disn}
Qiangeng Xu, Weiyue Wang, Duygu Ceylan, Radomir Mech, and Ulrich Neumann.
\newblock Disn: Deep implicit surface network for high-quality single-view 3d reconstruction.
\newblock \emph{Advances in neural information processing systems}, 32, 2019.

\bibitem[Xue et~al.(2023)Xue, Gao, Xing, Mart{\'\i}n-Mart{\'\i}n, Wu, Xiong, Xu, Niebles, and Savarese]{xue2023ulip}
Le~Xue, Mingfei Gao, Chen Xing, Roberto Mart{\'\i}n-Mart{\'\i}n, Jiajun Wu, Caiming Xiong, Ran Xu, Juan~Carlos Niebles, and Silvio Savarese.
\newblock Ulip: Learning a unified representation of language, images, and point clouds for 3d understanding.
\newblock In \emph{Proceedings of the IEEE/CVF Conference on Computer Vision and Pattern Recognition}, pp.\  1179--1189, 2023.

\bibitem[Yan et~al.(2016)Yan, Yang, Yumer, Guo, and Lee]{yan2016perspective}
Xinchen Yan, Jimei Yang, Ersin Yumer, Yijie Guo, and Honglak Lee.
\newblock Perspective transformer nets: Learning single-view 3d object reconstruction without 3d supervision.
\newblock \emph{Advances in neural information processing systems}, 29, 2016.

\bibitem[Yang et~al.(2023)Yang, Chen, Qian, Madaan, Iyengar, Fouhey, and Chai]{yang2023llmgrounder}
Jianing Yang, Xuweiyi Chen, Shengyi Qian, Nikhil Madaan, Madhavan Iyengar, David~F. Fouhey, and Joyce Chai.
\newblock Llm-grounder: Open-vocabulary 3d visual grounding with large language model as an agent, 2023.

\bibitem[Yu et~al.(2021)Yu, Ye, Tancik, and Kanazawa]{yu2021pixelnerf}
Alex Yu, Vickie Ye, Matthew Tancik, and Angjoo Kanazawa.
\newblock pixelnerf: Neural radiance fields from one or few images.
\newblock In \emph{Proceedings of the IEEE/CVF Conference on Computer Vision and Pattern Recognition}, pp.\  4578--4587, 2021.

\bibitem[Yu et~al.(2022)Yu, Wang, Vasudevan, Yeung, Seyedhosseini, and Wu]{yu2022coca}
Jiahui Yu, Zirui Wang, Vijay Vasudevan, Legg Yeung, Mojtaba Seyedhosseini, and Yonghui Wu.
\newblock Coca: Contrastive captioners are image-text foundation models.
\newblock \emph{arXiv preprint arXiv:2205.01917}, 2022.

\bibitem[Yu et~al.(2023)Yu, Xu, Zhang, Liu, Ye, Wu, Yan, Zhu, Xiong, Liang, et~al.]{yu2023mvimgnet}
Xianggang Yu, Mutian Xu, Yidan Zhang, Haolin Liu, Chongjie Ye, Yushuang Wu, Zizheng Yan, Chenming Zhu, Zhangyang Xiong, Tianyou Liang, et~al.
\newblock Mvimgnet: A large-scale dataset of multi-view images.
\newblock In \emph{Proceedings of the IEEE/CVF Conference on Computer Vision and Pattern Recognition}, pp.\  9150--9161, 2023.

\bibitem[Zeng et~al.(2023)Zeng, Jiang, Mao, Han, Ye, Huang, Yeung, Yang, Liang, and Xu]{zeng2023clip2}
Yihan Zeng, Chenhan Jiang, Jiageng Mao, Jianhua Han, Chaoqiang Ye, Qingqiu Huang, Dit-Yan Yeung, Zhen Yang, Xiaodan Liang, and Hang Xu.
\newblock Clip2: Contrastive language-image-point pretraining from real-world point cloud data.
\newblock In \emph{Proceedings of the IEEE/CVF Conference on Computer Vision and Pattern Recognition}, pp.\  15244--15253, 2023.

\bibitem[Zhang et~al.(2020)Zhang, Riegler, Snavely, and Koltun]{zhang2020nerf++}
Kai Zhang, Gernot Riegler, Noah Snavely, and Vladlen Koltun.
\newblock Nerf++: Analyzing and improving neural radiance fields.
\newblock \emph{arXiv preprint arXiv:2010.07492}, 2020.

\bibitem[Zhang et~al.(2022)Zhang, Kolkin, Bi, Luan, Xu, Shechtman, and Snavely]{zhang2022arf}
Kai Zhang, Nick Kolkin, Sai Bi, Fujun Luan, Zexiang Xu, Eli Shechtman, and Noah Snavely.
\newblock Arf: Artistic radiance fields.
\newblock In \emph{European Conference on Computer Vision}, pp.\  717--733. Springer, 2022.

\bibitem[Zhang et~al.(2018)Zhang, Isola, Efros, Shechtman, and Wang]{zhang2018lpips}
Richard Zhang, Phillip Isola, Alexei~A Efros, Eli Shechtman, and Oliver Wang.
\newblock The unreasonable effectiveness of deep features as a perceptual metric.
\newblock In \emph{CVPR}, 2018.

\end{thebibliography}
\bibliographystyle{iclr2024_conference}

\clearpage
\appendix
\section*{Appendices}

\section{Background of Model Components}
\label{sec:model_components}

\subsection{NeRF}

We adopt NeRF~\citep{mildenhall2021nerf}, specifically the compact triplane NeRF variant~\citep{chan2022eg3d}, as our 3D representation to predict in \ours{}. NeRF, when coupled with differentiable volume rendering, can be optimized with just image reconstruction losses. 

At the core of NeRF~\citep{mildenhall2021nerf} and its variants~\citep{chan2022eg3d,Chen2022ECCV,mueller2022instant,SunSC22} is a spatially-varying color (modeling appearance) and density (modeling geometry) field function. \footnote{To simplify the discussion, we ignore the view-dependent modeling in NeRF~\citep{mildenhall2021nerf}.} Given a 3D point $\mathbf{p}$, the color and density field $(\mathbf{u}, \sigma)$ can be written as: 
\begin{align}
 (\mathbf{u},\sigma) = \mathrm{MLP^\mathit{nerf}} (f_\theta(\mathbf{p})),
\end{align}
where the spatial encoding $f_\theta$ is used to facilitate the $\mathrm{MLP^\mathit{nerf}}$ to learn high-frequency signals. Different NeRF variants~\citep{chan2022eg3d,Chen2022ECCV,mueller2022instant,SunSC22} typically differ from each other in terms of the choice of the spatial encoding and the size of the MLP. In this work, we use the triplane spatial encoding function proposed by EG3D~\citep{chan2022eg3d}, because of its low tokenization complexity ($O(N^2)$ as opposed to a voxel grid's $O(N^3)$ complexity, where $N$ is spatial resolution).

Images are rendered from NeRF using volume rendering that's trivially differentiable. In detail, for each pixel to render, we cast a ray $\mathbf{r}$ through a NeRF, and use finite point samples $\mathbf{p}_i$ along the ray to compute the volume rendering integral to get the rendered color $\mathbf{u}(\mathbf{r})$:
\begin{align}
    \mathbf{u}(\mathbf{r}) & = \sum_{i} T_i (1 - \exp(- \sigma_i \delta_i)) \mathbf{u}_i, \\
    T_i &= \exp(- \sum_{j=1}^{i-1} \sigma_j \delta_j), 
\end{align}
where  $(\mathbf{u}_i,\sigma_i) = \mathrm{MLP}_\phi (f_\theta(\mathbf{p}_i))$ and $\delta_i$ is the distance between point $\mathbf{p}_i$ and $\mathbf{p}_{i+1}$.

\subsection{Transformer Layers}

In this subsection, we provide the details of the layers used in the transformer decoder~\citep{vaswani2017attention} as a background.
For the Vision Transformer encoder, please refer to the original DINO paper~\citep{caron2021dino} for implementation details.

\paragraph{Attention operator} 
Attention operator is an expressive neural operator which converts an input feature $x$ with condition to a sequence of other features $\{y_i\}$.
It first computes the attention score $\alpha_i$ by using the dot product between the input $x$ and each condition feature $y_i$.
An additional $\mathrm{softmax}$ is added after the dot products to normalize the weights to a summation of 1.
This attention score measures the relationship between input and conditions. 
Then the output is the weighted summation of the conditions $\{y_i\}$ with respect to the attention score $\alpha_i$.
\begin{align}
    \alpha_i &= \mathrm{softmax}_i \{x^\top y_i\} \\
    \mathrm{Attn}&(x; \{y_i\}_i) = \sum_i \alpha_i y_i
\end{align}

For some specific cases (\textit{e.g.}, in the transformer attention layer below), the attention operator wants to differentiate the vectors used in calculating the attention score and the vectors for final outputs.
Thus it will introduce another set of `value' vectors $\{z_i\}_i$, and treat the $\{y_i\}_i$ as corresponding `key' vectors.
Taking this into consideration, the formula would become
\begin{align}
    \alpha_i &= \mathrm{softmax}_i \{x^\top y_i\} \\
    \mathrm{Attn}&(x; \{y_i\}_i, \{z_i\}_i) = \sum_i \alpha_i z_i
\end{align}

\paragraph{Multi-head Attention}
The attention operator described above only attends to the condition features once to get the attention vector.
However, the actual attention might contain multiple modes. 
Thus, the multi-head attention \citep{vaswani2017attention} is proposed.
The multi-head attention is implemented by first splitting the input features into smaller queries.
\begin{align}
    [x^1, \ldots, x^\mathit{nh}] &= x 
\end{align}
where $\mathit{nh}$ is the number of heads. Meanwhile, $y_i$ and $z_i$ are split into $\{y^k_i\}_k$ and $\{z^k_i\}_k$ in a similar way. 
After that, the output of each head is computed independently and the final output is a concatenation of heads' outputs.
\begin{align}
    \mathit{out}^k &= \mathrm{Attn}(x^k; \{y^k_i\}_i, \{z^k_i\}_i) \\
    \mathrm{MultiHeadAttn}(x; \{y_i\}_i, \{z_i\}_i) &= [\mathit{out}^1, \ldots, \mathit{out}^\mathit{nh}]  
\end{align}

\begin{figure}[t]
  \centering
  \includegraphics[width=0.80\textwidth]{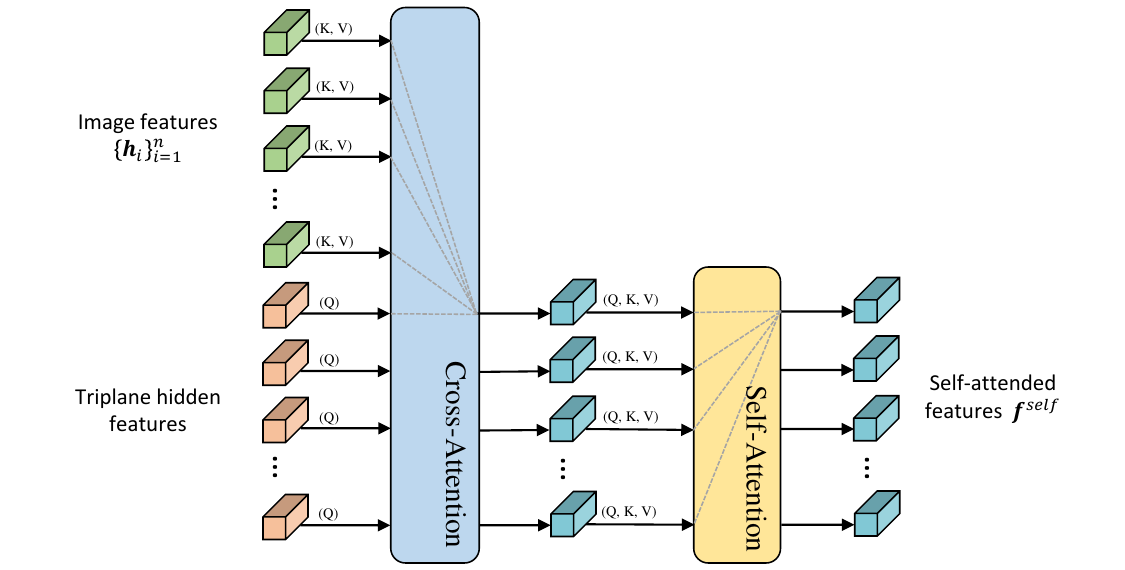}
  \vspace{-5pt}
  \caption{Visual illustration of the cross-attention and self-attention in \ours{}'s image-to-triplane decoder.}
  \label{fig:attention}
\end{figure}

\paragraph{Attention Layers in Transformer}
The detailed attention layers in transformer utilize the above multi-head attention with more linear layers. 
Here are the formulas for the self-attention layer (see the right yellow `Self-Attention' block in Fig.~\ref{fig:attention}). 
The layer first projects the input feature sequence $f=\{f_j\}_j$ to query $q$, key $k$, and value $v$ vectors with linear layers.
Then the multi-head attention is applied.
There is one more linear layer over the output.
We also follow the recent papers~\citep{chowdhery2022palm, touvron2023llama} to remove the bias terms in the attention layers.
\begin{align}
q_j &= W_\mathrm{q}f_j \\
k_i &= W_\mathrm{k}f_i \\
v_i &= W_\mathrm{v}f_i \\
o_j &= \mathrm{MultiHeadAttn}(q_j; \{k_i\}_i, \{v_i\}_i) \\
\mathrm{SelfAttn}(f_j; \{f_j\}_j) &= W_\mathrm{out} o_j \\
\end{align}
The cross-attention layer is defined similarly (see the left blue `Cross-Attention' block in Fig.~\ref{fig:attention}). 
The only difference to the self-attention layer is that the $W_\mathrm{k}$ and $W_\mathrm{v}$ is applied to the condition vectors (\textit{e.g.}, the image features $h$ in our example).

\paragraph{MLP layers in Transformer}
The Transformer model architecture applies the MLP layer (multi-layer perceptron) to do channel mixing (\textit{i.e.}, mix the information from different feature dimensions).
We follow the original transformer paper~\citep{vaswani2017attention} for the implementation.
The MLP layer contains two linear layers with a GELU~\citep{hendrycks2023gelu} activation in between.
The intermediate hidden dimension is 4 times of the model dimension.

\paragraph{Layer Normalization} 
We take the default LayerNorm (LN) implementation in PyTorch~\citep{paszke2019pytorch}. 
Besides the LN layers in $\mathrm{ModLN}$ as in Sec.~\ref{subsec:decoder}, we follow the Pre-LN architecture to also apply LN to the final output of transformers, \textit{e.g.}, the output of ViT and also the output of transformer decoder.

\paragraph{Positional Encoding} 
The positional embedding in ViT~\citep{dosovitskiy2020vit} is bilinearly upsampled from its original resolution ($14{\times}14$ for input $224{\times}224$) to match our higher input resolution ($32{\times}32$ for input $512{\times}512$).

\section{Training Setup}

We specify the training setup of our \ours{}. 
Apart from the information that we provided in Sec.~\ref{subsec:implement}, we apply a cosine schedule~\citep{loshchilov2016sgdr} with 3000 warm-up iterations.
We set the second beta parameter ($\beta_2$) of the AdamW optimizer~\citep{loshchilov2017adamw} to be 0.95.
We apply a gradient clipping of 1.0 and a weight decay of 0.05.
The weight decay are only applied on the weights that are not bias and not in the layer normalization layer.
We use BF16 precision in in the mixed precision training.
To save computational cost in training, we resize the reference novel views from $512{\times}512$ to a randomly chosen resolution between $128{\times}128$ and $384{\times}384$ and only ask the model to reconstruct a randomly selected $128{\times}128$ region. With this design, we can possibly increase the effective resolution of the model.

\section{Comparison with SoTA}
\label{sec:compare_sota}

We provide a quantitative comparison to the stat-of-the-art methods Point-E~\citep{nichol2022pointe}, Shap-E~\citep{jun2023shape}, and One-2-3-45~\citep{liu2023one2345}. Point-E trains an image-to-3D point cloud diffusion model, Shap-E encodes point clouds to latent representations and trains a diffusion model on the latents to generate parameters of a 3D implicit function, and One-2-3-45 reconstructs multi-view images generated with a 2D diffusion model. We randomly selected 100 objects from the Google Scanned Objects (GSO) dataset~\citep{downs2022googlescanned} and measured the novel view synthetic quality of 20 reference views (FID, CLIP-Similarity~\citep{radford2021clip}, PSNR, LPIPS~\citep{zhang2018lpips}) and the geometric quality (Chamfer Distance), as shown in the Table below. We can see that our LRM consistently outperforms previous approaches in all metrics.

\begin{table}[h]
  \caption{Comparison between LRM and state-of-the-art 3D generative models on Google Scanned Objects dataset (100 randomly selected objects and 20 reference views).}
  \begin{center}
  \resizebox{0.70\columnwidth}{!}{
  \begin{tabular}{l|cccccc}
    \hline \hline
    \multicolumn{1}{c|}{\multirow{2}{*}{Models}} & \multicolumn{5}{c}{GSO Evaluation} \\
    \cline{2-6} & 
    \multicolumn{1}{c}{FID$\downarrow$} & \multicolumn{1}{c}{CLIP-Similarity$\uparrow$} &
    \multicolumn{1}{c}{PSNR$\uparrow$} & \multicolumn{1}{c}{LPIPS$\downarrow$} & \multicolumn{1}{c}{Chamfer Distance$\downarrow$} \Tstrut\\
    \hline \hline
    Point-E & 123.70 & 0.741 & 15.60 & 0.308 & 0.099 \\ 
    Shap-E & 97.05 & 0.805 & 14.36 & 0.289 & 0.085 \\    
    One-2-3-45 & 139.24 & 0.713 & 12.42 & 0.448 & 0.123 \\
    \hline
    LRM (ours) & \textbf{31.44} & \textbf{0.902} & \textbf{19.60} & \textbf{0.163} & \textbf{0.053} \\
    \hline \hline
  \end{tabular}}
  \end{center}
  \label{tab:compare_sota}
\end{table}

We would like to discuss further the difference between LRM and the large-scale approaches Point-E and Shap-E. The models of Point-E and Shap-E contain hundreds of millions of learnable parameters and are trained with several million 3D assets (unknown data source and unknown computational cost from their papers). In terms of the network and dataset sizes, our LRM has 500 million learnable parameters, and it is trained on 1 million 3D data (publicly accessible), which does not show an advantage. In terms of the network architecture, Point-E, Shap-E, and LRM all use transformer-based models and apply cross-attention for inter-modality modeling (\textit{i.e.}, image-to-point cloud, point cloud+image-to-3D latents, and image-to-triplane, respectively).
We hypothesize it is the choice of very compact and expressive triplane representation together with an end-to-end trainable framework that enables the effective scaling of LRM and its adequate learning on large datasets (Objaverse and MvImgNet). Compared to the unstructured point cloud representation applied in Point-E and Shap-E, LRM applies the structured triplane representation that is aligned with the world frame, which naturally facilitates 2D-to-3D projection. It is also worth mentioning that Point-E uses 4K points (as tokens) and Shap-E uses 16K points (as tokens), but our LRM only uses $3{\times}32{\times}32{=}3072$ triplane tokens, which largely reduce the modeling complexity. Additionally, compared to the two-stage approach in Shape-E, which attempts to generate latents that can produce the parameters of implicit 3D functions through a diffusion model, our LRM directly maps 2D images to triplanes, which should be much more stable and efficient to learn.
Overall, we suggest that LRM is a more data-friendly and efficient model than Point-E and Shap-E.

\section{Analyses}
\label{sec:analysis}

We evaluate the effect of data, model hyper-parameters, and training methods on the performance of \ours{}, measuring by PSNR, CLIP-Similarity~\citep{radford2021clip}, SSIM~\citep{wang2004ssim} and LPIPS~\citep{zhang2018lpips} of the rendered novel views. Note that due to the large training cost of our final model, the following analytic experiments use a much smaller version of \ours{} model as the baseline (indicated by orange shaded rows in the tables). Specifically, we scale down the image-to-triplane decoder to 12 cross-attention layers, change the input image resolution to 256, triplane latent dimension to 32, rendering resolution in training to 64, and use 96 samples per ray for rendering $64{\times}64$ images for supervision. We only train each model on 32 NVIDIA A100 GPUs for 15 epochs, and the resulting difference can be seen in Table~\ref{tab:final_base}.
We are aware that some observations might change if we scale up the model, but most of the conclusions should be general and consistent.

\begin{table}[h]
  \caption{Comparison between the final model and the baseline for analysis.}
  \begin{center}
  \resizebox{0.50\columnwidth}{!}{
  \begin{tabular}{l|ccccc}
    \hline \hline
    \multicolumn{1}{c|}{\multirow{2}{*}{Models}} & \multicolumn{4}{c}{Unseen Evaluation} \\
    \cline{2-5} & 
    \multicolumn{1}{c}{PSNR$\uparrow$} & \multicolumn{1}{c}{CLIP-Similarity$\uparrow$} &
    \multicolumn{1}{c}{SSIM$\uparrow$} & \multicolumn{1}{c}{LPIPS$\downarrow$} \Tstrut\\
    \hline \hline
    Final & \textbf{20.1} & \textbf{91.0} & \textbf{79.7} & \textbf{16.0} \\ 
    \rowcolor{orange!50} Baseline & 19.0 & 87.8 & 77.4 & 19.1  \\
    \hline \hline
  \end{tabular}}
  \end{center}
  \label{tab:final_base}
\end{table}

\subsection{Synthetic vs. Real Data}
Table~\ref{tab:abl_datasets} compares the influence of using synthetic 3D data from the Objaverse~\citep{deitke2023objaverse} and real video data from the MvImgNet~\citep{yu2023mvimgnet} in training. Results show that removing real data causes an obvious drop for all the metrics, despite the fact our synthetic 3D dataset contains $3{\times}$ more shapes than MvImgNet. One potential reason is that the real data have much more variation in the lighting, the size of the target, and the camera poses, which effectively benefits the learning. Future work could augment the rendering of synthetic shapes to adequately utilize those abundant data. Nevertheless, combining the two datasets leads to substantially better results than training on any one of them alone.

\begin{table}[h!]
  \caption{Influence of training datasets.}
  \begin{center}
  \resizebox{0.57\columnwidth}{!}{
  \begin{tabular}{l|cccc}
    \hline \hline
    \multicolumn{1}{c|}{\multirow{2}{*}{Data}} & \multicolumn{4}{c}{Unseen Evaluation} \\
    \cline{2-5} & 
    \multicolumn{1}{c}{PSNR$\uparrow$} & \multicolumn{1}{c}{CLIP-Similarity$\uparrow$} &
    \multicolumn{1}{c}{SSIM$\uparrow$} & \multicolumn{1}{c}{LPIPS$\downarrow$} \Tstrut\\
    \hline \hline
    Synthetic (Objaverse) & 15.5 & 84.7 & 70.3 & 29.3 \\ 
    Real (MvImgNet) & 17.5 & 85.7 & 75.7 & 22.0 \\
    \hline
    \rowcolor{orange!50} Synthetic+Real & \textbf{19.0} & \textbf{87.8} & \textbf{77.4} & \textbf{19.1} \\ 
    \hline \hline
  \end{tabular}}
  \end{center}
  \label{tab:abl_datasets}
\end{table}

\subsection{Number of Views in Training Data}
In Table~\ref{tab:abl_trainviews}, we conduct experiments with all data but limit the number of training views per shape.
For example, for \textit{Train Views$=$8}, we use only a random subset of 8 views per shape and keep randomly sampling 4 views from the above subset at each training step.
The results show that more views can lead to better results, possibly because of more diverse data. While the growth is saturated at 16 views, adding more views does not lead to worse results.

\begin{table}[h!]
  \caption{Effect of the number of different views per shape in training. 32+ indicates some video data in MvImgNet contain more than 32 views per shape, which we apply all of them in training.}
  \begin{center}
  \resizebox{0.5\columnwidth}{!}{
  \begin{tabular}{l|cccc}
    \hline \hline
    \multicolumn{1}{c|}{\multirow{2}{*}{Train Views}} & \multicolumn{4}{c}{Unseen Evaluation} \\
    \cline{2-5} & 
    \multicolumn{1}{c}{PSNR$\uparrow$} & \multicolumn{1}{c}{CLIP-Similarity$\uparrow$} &
    \multicolumn{1}{c}{SSIM$\uparrow$} & \multicolumn{1}{c}{LPIPS$\downarrow$} \Tstrut\\
    \hline \hline
    4  & 18.8 & 86.7 & 77.5 & 19.8 \\ 
    8  & 18.9 & 87.3 & 77.5 & 19.4 \\ 
    16 & \textbf{19.1} & \textbf{87.9} & \textbf{77.6} & \textbf{19.0} \\ 
    \hline
    \rowcolor{orange!50} 32+ &  19.0 & 87.8 & 77.4 & 19.1 \\
    \hline \hline
  \end{tabular}}
  \end{center}
  \label{tab:abl_trainviews}
\end{table}

\subsection{Model Hyper-parameters}

Table~\ref{tab:abl_crossattn_num} presents the results of having a different number of cross-attention layers in the image-to-triplane decoder. There is a slight trend indicating that the scores can be improved by having a deeper model, especially for the latent semantic and perceptual similarity measurements CLIP and LPIPS, implying that the network models better representations for reconstructing higher-quality images.

We also evaluate the influence of the number of MLP layers in NeRF (Table~\ref{tab:abl_mlpnerf}). Results show that it is unnecessary to have a very large network, and there seems to be a sweet spot around two to four layers. This observation is consistent with EG3D~\citep{chan2022eg3d} where the information of shapes is encoded by the triplane and such MLP is only a shallow model for projecting triplane features to color and density.

As shown in Table~\ref{tab:abl_triplane_size}, we found that increasing the triplane resolution leads to better image quality. Note that, in this experiment, we only use a deconvolution layer to upsample the $32{\times}32{\times}32$ triplane produced by LRM's decoder, whereas we suspect a large improvement could be seen by increasing the quantity of input spatial-positional embeddings to query more fine-grained image details. However, such an approach will dramatically increase the computational cost, we leave this exploration to future research.

\begin{table}[h!]
  \caption{Effect of the number of cross-attention layers in image-to-triplane decoder.}
  \begin{center}
  \resizebox{0.5\columnwidth}{!}{
  \begin{tabular}{l|cccc}
    \hline \hline
    \multicolumn{1}{c|}{\multirow{2}{*}{\begin{tabular}{@{}c@{}}CrossAttn \\ Layers\end{tabular}}} & \multicolumn{4}{c}{Unseen Evaluation} \\
    \cline{2-5} & 
    \multicolumn{1}{c}{PSNR$\uparrow$} & \multicolumn{1}{c}{CLIP-Similarity$\uparrow$} &
    \multicolumn{1}{c}{SSIM$\uparrow$} & \multicolumn{1}{c}{LPIPS$\downarrow$}\Tstrut\\
    \hline \hline
    6  & 19.0 & 87.7 & \textbf{77.6} & 19.1 \\
    \rowcolor{orange!50} 16 & 19.0 & 87.8 & 77.4 & 19.1  \\
    24 & \textbf{19.1} & \textbf{88.0} & \textbf{77.6} & \textbf{18.9} \\
    \hline \hline
  \end{tabular}}
  \end{center}
  \label{tab:abl_crossattn_num}
\end{table}

\begin{table}[h!]
  \caption{Effect of the number of MLP layers in NeRF.}
  \begin{center}
  \resizebox{0.5\columnwidth}{!}{
  \begin{tabular}{l|cccc}
    \hline \hline
    \multicolumn{1}{c|}{\multirow{2}{*}{\begin{tabular}{@{}c@{}}NeRF MLP \\ Layers\end{tabular}}} & \multicolumn{4}{c}{Unseen Evaluation} \\
    \cline{2-5} & 
    \multicolumn{1}{c}{PSNR$\uparrow$} & \multicolumn{1}{c}{CLIP-Similarity$\uparrow$} &
    \multicolumn{1}{c}{SSIM$\uparrow$} & \multicolumn{1}{c}{LPIPS$\downarrow$} \Tstrut\\
    \hline \hline
    2  & \textbf{19.2} & 87.7 & \textbf{77.8} & \textbf{18.9} \\    
    6  & 19.1 & \textbf{88.0} & 77.6 & 19.0 \\
    \rowcolor{orange!50} 12 & 19.0 & 87.8 & 77.4 & 19.1 \\
    14 & 19.1 & 87.2 & 77.6 & 19.0 \\
    \hline \hline
  \end{tabular}}
  \end{center}
  \label{tab:abl_mlpnerf}
\end{table}

\begin{table}[h!]
  \caption{Effect of the resolution of triplane. For \textit{64up} and \textit{128up}, we apply additional $2{\times}2$ and $4{\times}4$ deconvolution layers, respectively, to upsample a \textit{Res.} 32 triplane.}
  \begin{center}
  \resizebox{0.5\columnwidth}{!}{
  \begin{tabular}{l|cccc}
    \hline \hline
    \multicolumn{1}{c|}{\multirow{2}{*}{Triplane Res.}} & \multicolumn{4}{c}{Unseen Evaluation} \\
    \cline{2-5} & 
    \multicolumn{1}{c}{PSNR$\uparrow$} & \multicolumn{1}{c}{CLIP-Similarity$\uparrow$} &
    \multicolumn{1}{c}{SSIM$\uparrow$} & \multicolumn{1}{c}{LPIPS$\downarrow$}\Tstrut\\
    \hline \hline
    32   & 18.9 & 86.3 & 77.2 & 19.7 \\
    \hline
    \rowcolor{orange!50} 64up & \textbf{19.0} & 87.8 & 77.4 & 19.1 \\
    128up  & \textbf{19.0} & \textbf{88.3} & \textbf{77.5} & \textbf{19.0} \\
    \hline \hline
  \end{tabular}}
  \end{center}
  \label{tab:abl_triplane_size}
\end{table}

\subsection{Camera Pose}

As we have discussed in the Main Paper, normalizing camera poses in training has a huge impact on the generalization of input views. We can see from Table~\ref{tab:abl_camera_pose} that when no modification is applied (\textit{None}), \ours{} produces the worst results. Augmenting camera poses with a \textit{Random} rotation greatly improves the results since the model learns a more general image-to-triplane projection via decoupled views and camera poses. However, such unconstrained projection is very difficult to learn. We therefore \textit{Normalized} all camera poses so that all images are projected onto the triplane from the same direction, allowing the model to adequately learn and utilize the cross-shape prior for reconstruction.

\begin{table}[h!]
  \caption{Effect of camera pose normalization.}
  \begin{center}
  \resizebox{0.5\columnwidth}{!}{
  \begin{tabular}{l|cccc}
    \hline \hline
    \multicolumn{1}{c|}{\multirow{2}{*}{Camera Pose}} & \multicolumn{4}{c}{Unseen Evaluation} \\
    \cline{2-5} & 
    \multicolumn{1}{c}{PSNR$\uparrow$} & \multicolumn{1}{c}{CLIP-Similarity$\uparrow$} &
    \multicolumn{1}{c}{SSIM$\uparrow$} & \multicolumn{1}{c}{LPIPS$\downarrow$}\Tstrut\\
    \hline \hline
    None   & 15.3 & 83.4 & 70.1 & 28.9 \\
    Random & 18.0 & 85.6 & 75.7 & 21.1 \\
    \hline
    \rowcolor{orange!50} Normalized & \textbf{19.0} & \textbf{87.8} & \textbf{77.4} & \textbf{19.1} \\
    \hline \hline
  \end{tabular}}
  \end{center}
  \label{tab:abl_camera_pose}
\end{table}

\subsection{Image Quantity and Resolution}

Table~\ref{tab:abl_sideviews} and Table~\ref{tab:abl_renderres} study the influence of the number of side views supervision for each sample and the effect of image rendering resolution in training. Results indicate that as the quantity of side views increases, the reconstructed image quality improves. Having more views allows the model to better correlate the appearance and geometry of different parts of the same shape, and facilitates inferring multi-view consistent results. Moreover, using a higher rendering resolution of images in training largely improves the results, as the model is encouraged to learn more high-frequency details.

\begin{table}[!h]
  \caption{Influence of the number of side views applied for each training sample.}
  \begin{center}
  \resizebox{0.5\columnwidth}{!}{
  \begin{tabular}{l|cccc}
    \hline \hline
    \multicolumn{1}{c|}{\multirow{2}{*}{Side Views}} & \multicolumn{4}{c}{Unseen Evaluation} \\
    \cline{2-5} & 
    \multicolumn{1}{c}{PSNR$\uparrow$} & \multicolumn{1}{c}{CLIP-Similarity$\uparrow$} &
    \multicolumn{1}{c}{SSIM$\uparrow$} & \multicolumn{1}{c}{LPIPS$\downarrow$} \Tstrut\\
    \hline \hline
    1 & 18.7 & 87.7 & 77.2 & 19.7 \\    
    2 & 18.7 & 87.5 & 77.2 & 19.6 \\
    \rowcolor{orange!50} 3 & 19.0 & \textbf{87.8} & 77.4 & 19.1 \\
    4 & \textbf{19.1} & \textbf{87.8} & \textbf{77.6} & \textbf{18.9} \\
    \hline \hline
  \end{tabular}}
  \end{center}
  \label{tab:abl_sideviews}
\end{table}

\begin{table}[!h]
  \caption{Influence of the rendering resolution of images in training.}
  \begin{center}
  \resizebox{0.5\columnwidth}{!}{
  \begin{tabular}{l|cccc}
    \hline \hline
    \multicolumn{1}{c|}{\multirow{2}{*}{Render Res.}} & \multicolumn{4}{c}{Unseen Evaluation} \\
    \cline{2-5} & 
    \multicolumn{1}{c}{PSNR$\uparrow$} & \multicolumn{1}{c}{CLIP-Similarity$\uparrow$} &
    \multicolumn{1}{c}{SSIM$\uparrow$} & \multicolumn{1}{c}{LPIPS$\downarrow$} \Tstrut\\
    \hline \hline
    32  & 18.8 & 86.3 & 77.0 & 20.1 \\ 
    \rowcolor{orange!50} 64 & 19.0 & 87.8 & 77.4 & 19.1 \\
    128 & \textbf{19.4} & \textbf{89.0} & \textbf{78.3} & \textbf{18.0} \\
    \hline \hline
  \end{tabular}}
  \end{center}
  \label{tab:abl_renderres}
\end{table}

\subsection{LPIPS Loss}

 Lastly, we found that our LPIPS objective~\citep{zhang2018lpips} has a huge impact on the results. Removing it from training will decrease the CLIP-Similarity, SSIM, and LPIPS scores to 74.7, 76.4, and 29.4, respectively.

\section{Visualizations}
\label{sec:appendix_visualize}

We present more visualizations of the reconstructed 3D shapes in the following pages. The input images include photos captured by our phone camera, images from Objaverse~\citep{deitke2023objaverse}, MvImgNet~\citep{yu2023mvimgnet}, ImageNet~\citep{deng2009imagenet}, Google Scanned Objects~\citep{downs2022googlescanned}, Amazon Berkeley Objects~\citep{collins2022abo}, and images generated by the Adobe Firefly\footnote{Adobe Firefly, a text-to-image generation tool: \href{https://firefly.adobe.com}{https://firefly.adobe.com}.}. 
We implement a heuristic function to pre-process the camera-captured images, generated images, and images from MvImgNet and ImageNet. The function removes the image background with an off-the-shelf package\footnote{Rembg package, a tool to remove image background: \href{https://pypi.org/project/rembg}{https://pypi.org/project/rembg}}, followed by cropping out the target object, rescaling the target to a suitable size and centering the target on a square white figure.
All input images are never seen by the model in training. Please visit our project webpage \textcolor{red}{\tt\small\url{https://yiconghong.me/LRM/}} for video demonstrations and interactable 3D meshes.

\begin{figure}[!h]
  \centering
  \vspace{20pt}
  \includegraphics[width=\textwidth]{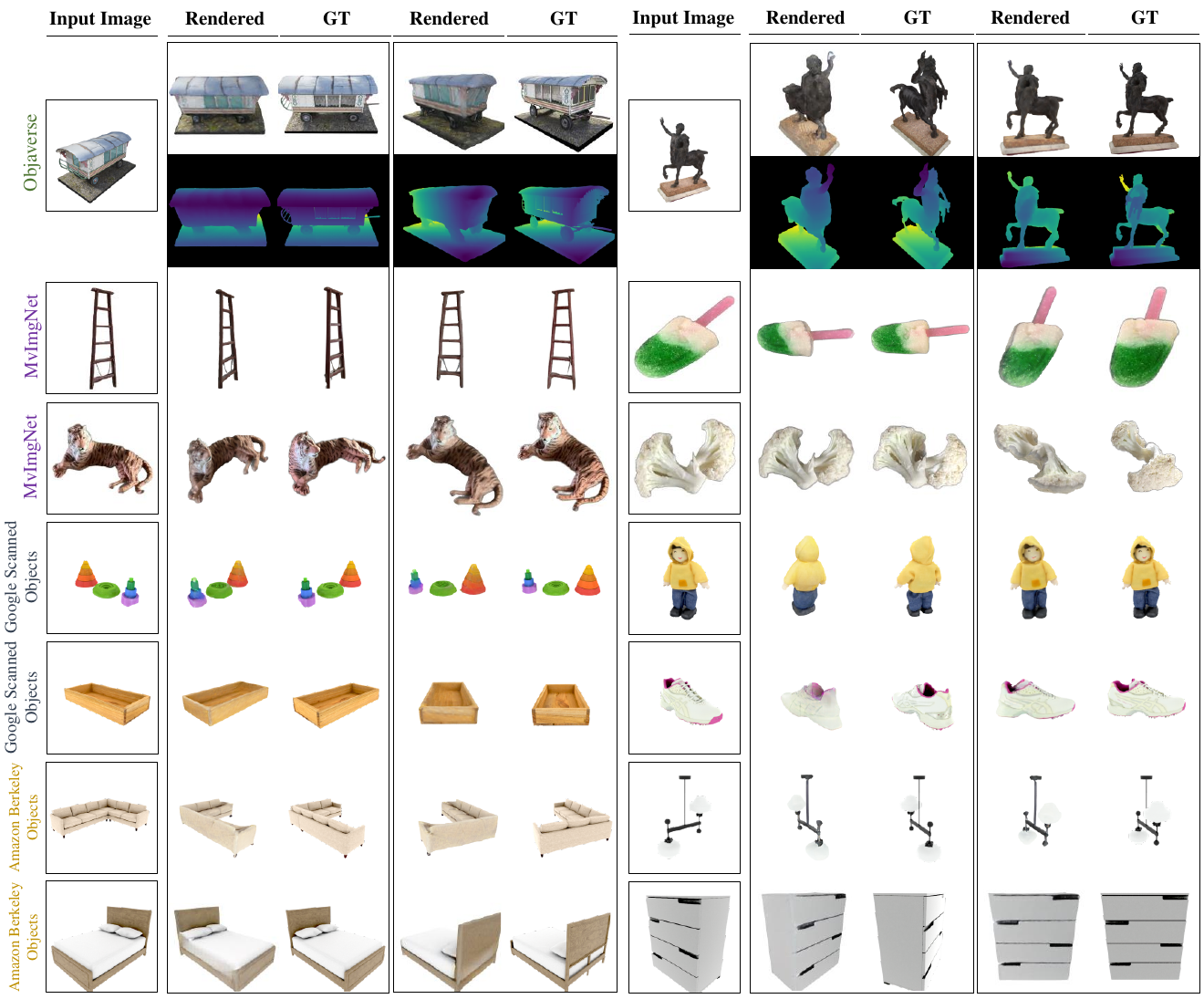}
  \vspace{-10pt}
  \caption{Comparison between \ours{} rendered novel views and the ground truth images (GT). None of the images are observed by the model during training. The GT depth images of Objaverse are rendered from the 3D models. Please zoom in for clearer visualization.}
  \label{fig:visualize_self_A3}
\end{figure}

 \begin{figure}[p]
  \centering
  \includegraphics[width=\textwidth]{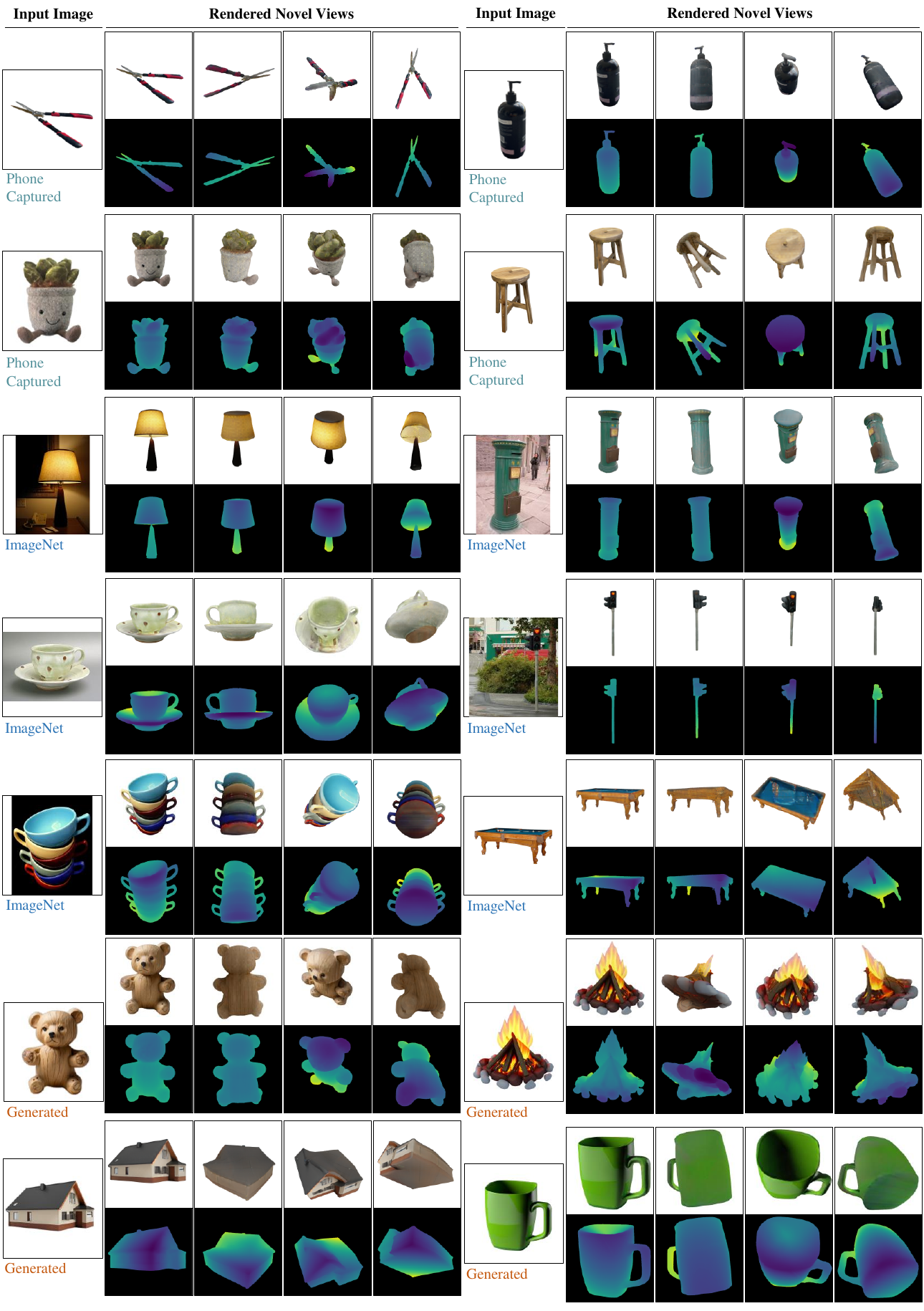}
  \vspace{-10pt}
  \caption{Rendered novel views (RGB and Depth) of shapes reconstructed by our \ours{} from single images. None of the images are observed by the model during training. Generated images are created by the Adobe Firefly. Please zoom in for clearer visualization.}
  \label{fig:visualize_self_A2}
\end{figure}

\end{document}